\documentclass{article} % For LaTeX2e
\usepackage[final]{colm2026_conference}

\usepackage{amsmath,amsfonts,bm}

\def\eqref#1{equation~\ref{#1}}
\def\1{\bm{1}}

\DeclareMathAlphabet{\mathsfit}{\encodingdefault}{\sfdefault}{m}{sl}
\SetMathAlphabet{\mathsfit}{bold}{\encodingdefault}{\sfdefault}{bx}{n}

\usepackage{microtype}
\usepackage{hyperref}
\usepackage{url}
\usepackage{booktabs}
\usepackage{graphicx}
\usepackage{tabularx}
\usepackage{multirow}
\usepackage{amsfonts}
\usepackage{amsmath}
\usepackage{amssymb}
\usepackage{mathtools}
\usepackage{amsthm}
\usepackage{xspace}
\usepackage{lipsum}
\usepackage{tcolorbox}
\usepackage{adjustbox}
\usepackage{makecell}
\usepackage{enumitem}
\setlist[enumerate]{topsep=0pt, partopsep=0pt, parsep=0pt, itemsep=0pt}
\setlist[itemize]{topsep=0pt, partopsep=0pt, parsep=0pt, itemsep=0pt}
\usepackage{pgfplots}
\pgfplotsset{compat=1.18}
\usepgfplotslibrary{groupplots}
\usepackage{listings}
\usepackage{subcaption}
\usepackage{pgf-pie}
\usepackage[capitalize,noabbrev]{cleveref}
\usepackage{lineno}
\usepackage[T1]{fontenc}

\definecolor{darkblue}{rgb}{0, 0, 0.5}
\hypersetup{colorlinks=true, citecolor=darkblue, linkcolor=darkblue, urlcolor=darkblue}

\lstdefinestyle{mypython}{
  language=Python,
  basicstyle=\ttfamily\footnotesize, 
  keywordstyle=\bfseries\color{blue},
  stringstyle=\color{orange},
  commentstyle=\color{gray},
  numbers=none,
  numberstyle=\tiny\color{gray},
  stepnumber=1,
  numbersep=5pt,
  frame=single,
  breaklines=true,           
  breakatwhitespace=true,    
  columns=fullflexible,      
  keepspaces=true,
  showstringspaces=false,
  tabsize=2,
  captionpos=b,
  xleftmargin=0pt,           
  xrightmargin=0pt,
  linewidth=\columnwidth     
}

\newcommand*\samethanks[1][\value{footnote}]{\footnotemark[#1]}
\newcommand{\np}{\texttt{NaturalPlan}\xspace}
\newcommand{\zl}{\texttt{ZebraLogic}\xspace}
\newcommand{\las}{$\textit{LLM-as-solver}$\xspace}
\newcommand{\laf}{$\textit{LLM-as-formalizer}$\xspace}

\newcommand{\drexel}{%
  \hspace{1pt}
  \begingroup\normalfont
  \includegraphics[height=1.3\fontcharht\font`\B]{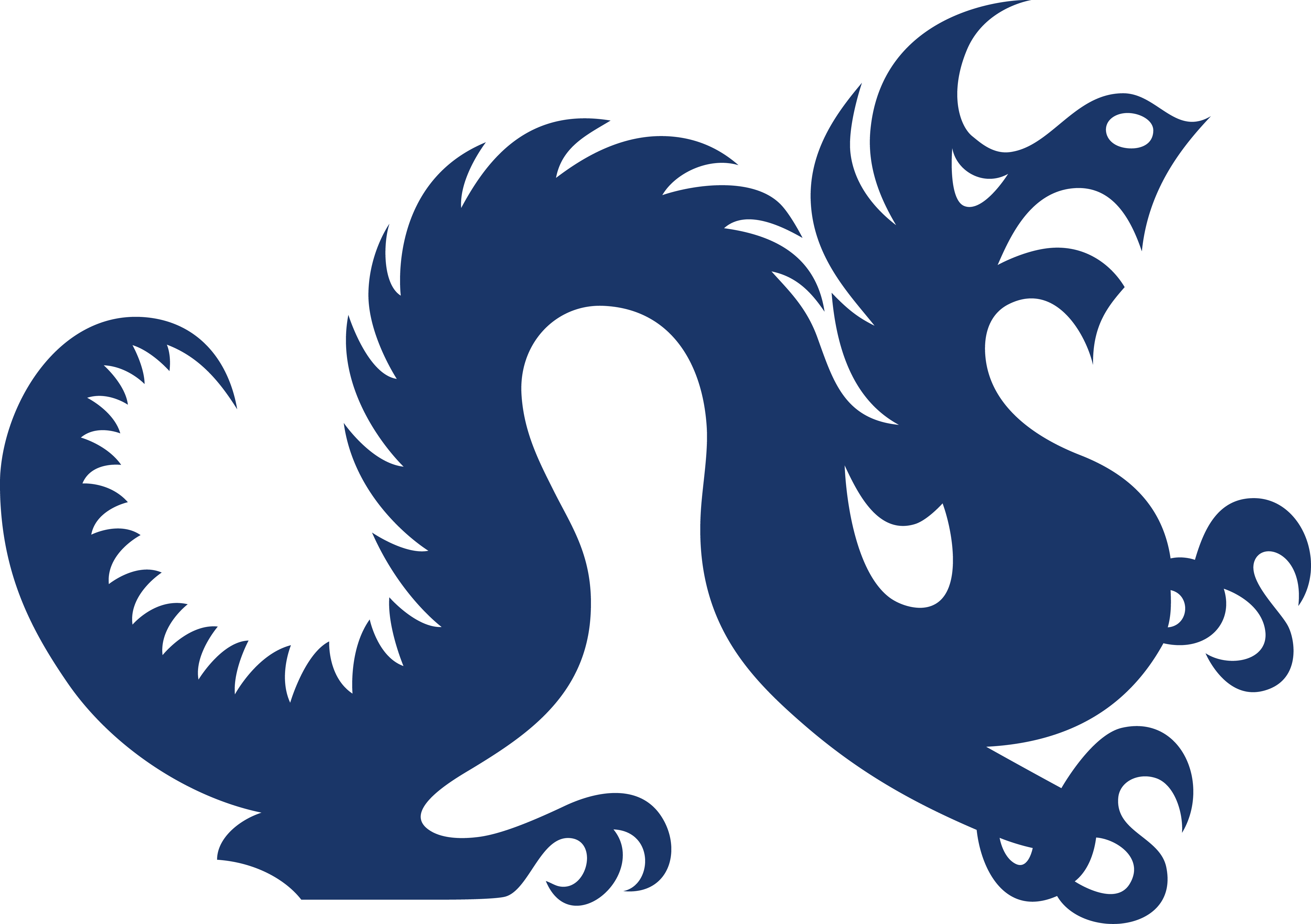}%
  \endgroup
  \hspace{1pt}
}
\newcommand{\aitwo}{%
  \hspace{1pt}%
  \begingroup\normalfont
  \includegraphics[height=1.3\fontcharht\font`\B]{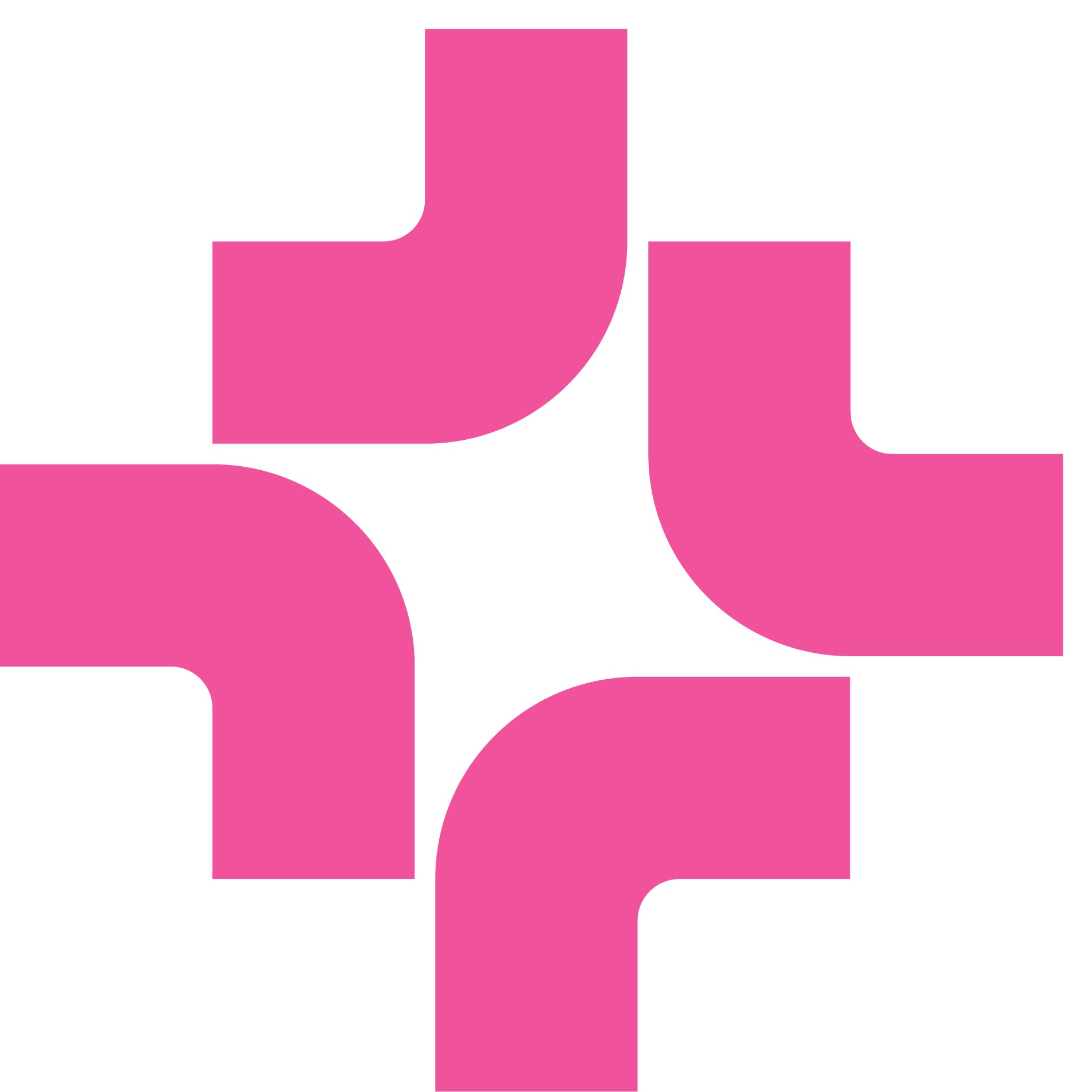}%
  \endgroup
  \hspace{1pt}%
}
\newcommand{\penn}{%
  \hspace{1pt}%
  \begingroup\normalfont
  \includegraphics[height=1.3\fontcharht\font`\B]{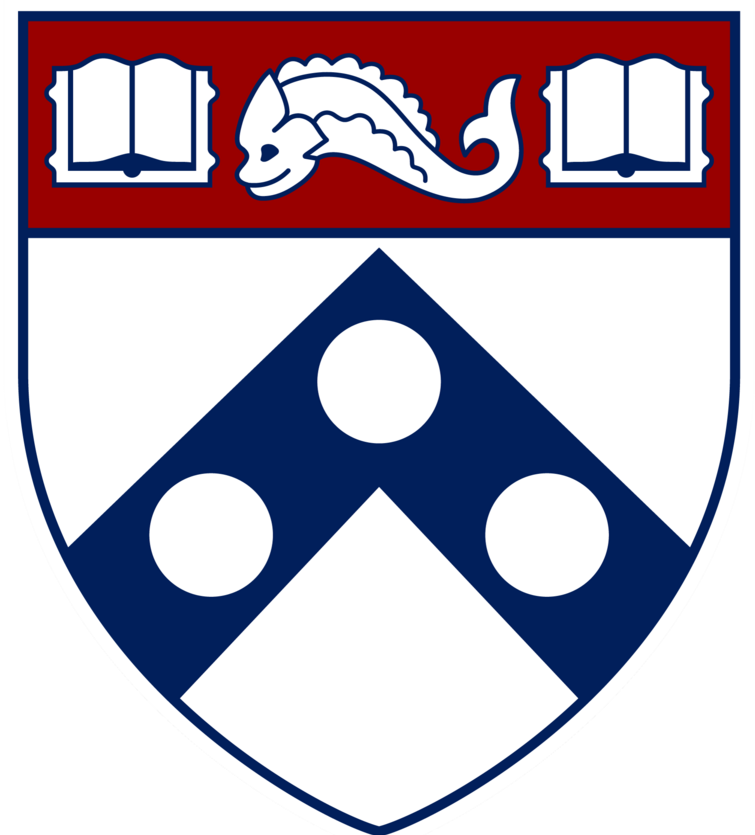}%
  \endgroup
  \hspace{1pt}%
}

\theoremstyle{plain}

\theoremstyle{definition}

\theoremstyle{remark}

\title{A Reality Check of Language Models as Formalizers \\on Constraint Satisfaction Problems}

\author{Rikhil Amonkar\thanks{Equal contribution.} \drexel \enspace Ceyhun Efe Kayan\samethanks\ \drexel \enspace Qimei Lai \penn \\ \textbf{Ronan Le Bras} \aitwo \enspace \textbf{Li Zhang} \drexel \\
  \drexel Drexel University \enspace \penn University of Pennsylvania \enspace \aitwo Allen Institute for AI \hspace{4pt} \\
  {\texttt{\{rikhil.amonkar|harry.zhang}@drexel.edu\}}
}

\begin{document}

\ifcolmsubmission
\linenumbers
\fi

\maketitle

\begin{abstract}
Recent work shows superior performance when using large language models (LLMs) as formalizers instead of as end-to-end solvers for symbolic reasoning problems. Given the problem description, the LLM generates a formal program that derives a solution via an external solver. We systematically investigate the formalization capability of LLMs on real-life constraint satisfaction problems on 4 benchmarks, 6 LLMs, and 2 types of formal languages. We show that  LLM-as-formalizer by no means trivializes the problem but underperforms LLM-as-solver in 15 out of 24 model-dataset combinations, despite the former's verifiability and interpretability. Although the formalization space is magnitudes smaller than the search space, our scaling analysis shows that LLM-as-formalizer still drastically degrades as problem complexity increases similar to LLM-as-solver. To better understand this limitation, we observe excessive, solver-like reasoning tokens that sometimes lead to hard-coded solutions, highlighting a key challenge for improving LLM-based formalization.\footnote{Code: \url{github.com/rikhil-amonkar/llm-csp}}
\end{abstract}

\section{Introduction}

Considerable efforts have been dedicated to improving the ability of large language models (LLMs) to solve logical reasoning tasks \citep{saxton2018analysing,10.5555/3491440.3491977}, primarily in an end-to-end manner involving intermediate chain-of-thought tokens \citep{10.5555/3600270.3602070,10.5555/3600270.3601883}.
Despite their strong performance, these \las methods have been shown to lack faithfulness, verifiability, and formal guarantee \citep{lyu-etal-2023-faithful}.
To bridge this gap, an emerging line of neuro-symbolic methods use LLMs to translate the natural language problem description into a formal program, from which a solution can be derived using an external solver \citep{10.5555/3618408.3618843,pan-etal-2023-logic,han-etal-2024-folio}.
Despite the desirable features above, the reported triumphant performance of formalization predates the advance of large reasoning models (LRMs) that scales reasoning tokens during inference time \citep{muennighoff2025s1simpletesttimescaling,guo2025deepseek}. Most recent work shows the performance comparison between two methodologies remains unclear and debated \citep{huang-zhang-2025-limit,kagitha2025addressingchallengesplanninglanguage}, hindering real-life application.

We systematically compare these two methodologies on constraint satisfaction problems (CSPs) following existing work that form the backbone of many real-world planning and scheduling systems, including calendar scheduling, trip planning, and meeting planning, in addition to a logic grid puzzle.
On these 4 domains, we evaluate 6 state-of-the-art LLMs, including 4 LRMs (DeepSeek-R1, Qwen3-32B, o3-mini-high, GPT-5) and 2 non-reasoning counterparts (DeepSeek-V3, Qwen2.5-32B), both as solvers and as formalizers.
To ensure maximal generalizability and fairness of comparison, we consider only one-shot settings with a minimal instruction to generate a solution or a code, but additionally provide formalizers with revision attempts. 
We consider two formalization targets: free-form Python and specific code to interface a Satisfiability Modulo Theories (SMT) solver.

\begin{figure}
    \centering
    \includegraphics[width=\columnwidth]{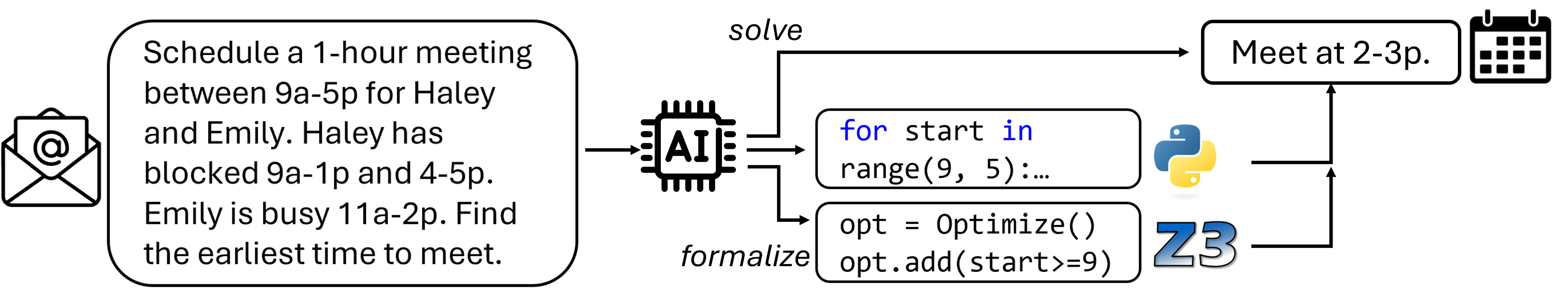}
    \caption{An example of the output for \las (top) and \laf with two types of formalism (middle and bottom) in a real-life constraint satisfaction task, calendar scheduling.}
    \label{fig:example}
\end{figure}

While existing work established the superiority of \laf, we instead show that it underperforms \las in 12 out of 16 LRM-dataset combinations and 4 out of 8 non-reasoning-LLM-dataset combinations. While many expect \laf to scale with problem complexity since the formalization space is magnitudes smaller than the search space, our analysis shows that its performance still drastically degrades as problem complexity increases just as \las. We perform a fine-grained error analysis and attribute the anti-climatic performance of \laf largely to missing constraints and erroneous logic. Despite the comparative token efficiency of \laf for LRMs, we observe surprising, excessive, solver-like reasoning tokens that sometime lead to hard-coded solutions, likely an artifact of pre-training. Our work suggests multiple future directions spanning performance, robustness, and efficiency of neurosymbolic auto-formalization, which promises verifiability and interpretability.

\section{Related Work}

\subsection{\las}

\las, also known as informal reasoning or end-to-end reasoning, is a paradigm when one or more LLMs generate optional intermediate tokens and eventually the solution to a logical reasoning problem.
As early problem solving ability comes from training, many major LLMs have been trained or evaluated as a solver \citep{rajani-etal-2019-explain,10.5555/3495724.3495883}, while limited to low-complexity, common-sense problems.
The introduction of chain-of-thought reasoning made \las feasible in complex tasks such as mathematics and logic puzzles \citep{10.5555/3600270.3602070,10.5555/3600270.3601883,10.5555/3692070.3693867,feng2023towards}.
Follow-up techniques to further improve \las based on chain-of-thought included self-verification \citep{weng-etal-2023-large}, self-refine \citep{madaan2023selfrefine}, tree-of-thought \citep{10.5555/3666122.3666639}, self-consistency \citep{wang2023selfconsistency}, etc.

Despite strong performance, \las has been shown to lack faithfulness, verifiability, and formal guarantees \citep{lyu-etal-2023-faithful,10.5555/3666122.3669397,stechly2025on}.
To bridge these gaps, neuro-symbolic methods were introduced to combine \las with formal tools \citep{10207581,saparov2023language}.

\subsection{\laf}
\label{sec:relwork-laf}

\laf is a subset of the above neuro-symbolic methods where LLMs generate neither the solution nor the chain-of-thought towards the solution, but rather translate the natural language problem description to an executable program based on LLM's ability to generate code.
This methodology has been reported to greatly outperform \las in classical planning \citep{liu2023llm+,xie2023translating,hao2025planning}, constraint satisfaction \citep{berman2024solvingzebrapuzzlesusing,kesseli2025logicpybridginggapllms}, mathematics \citep{10.5555/3600270.3602614,10.5555/3666122.3667066,jiang2023draft}, and general logical reasoning tasks \citep{10.5555/3666122.3668096, 10.5555/3618408.3618843,pan-etal-2023-logic,han-etal-2024-folio}.
Despite the overwhelming report of its success, it has also been shown to be unstable, dependent on the choice of formal language and solver \citep{matthew-lam-etal-2024-closer,beiser2025intermediatelanguagesmatterformal}.

\subsection{\laf vs. \las}

For effective application, research like ours that systematically compares \laf and \las is crucial but lacking, especially with the introduction of LRMs, which reported dramatically improved performance on algorithmic tasks where \las used to fail \citep{muennighoff2025s1simpletesttimescaling,guo2025deepseek}.
The closest work to ours is \citet{chen2025steering}, which did not consider any open-source model or inference-time scaling LRMs, whereas open-source LRMs are our focus.
Moreover, their work is an initial investigation where conclusions are surface-level (formalizing is hard) and inconclusive (neither is optimal), while ours is a deep dive resulting in an array of fine-grained, relatively definitive, and somewhat counterintuitive conclusions.
The other work that considered both \laf and \las for LRMs is \citet{kagitha2025addressingchallengesplanninglanguage}, which only focuses on classical planning tasks, similarly lacking decisive findings.

\begin{table}[t!]
    \centering
    \small
    \resizebox{\columnwidth}{!}{%
    \begin{tabular}{lllll}
    \toprule
         & Calendar Scheduling & Trip Planning  & Meeting Planning & Zebra Logic \\ \midrule
        $X_i$ & start hour $s$, end hour $e$ & day $d$, city $c$ & person $p$, start $s$, duration $d$ & person $p$, attribute $a$\\ \\
        $D_i$  & \makecell[tl]{$\{(s,e)\mid s,e \in [9,17]$\\$e>s\}$} & \makecell[tl]{$\{(d,c)\mid d\in [1,\text{len}],$\\$c\in C\}$} & \makecell[tl]{$\{(p,s,d)\mid p\in P,s\in [9,24],$\\$d\in [1,12]\}$} & $\{(p,a)\mid p \in P, a \in A\}$\\ \\
        $C_j$ & \makecell[tl]{Unavailable time range\\Meeting duration} & \makecell[tl]{Total number of days\\Allowed direction of travel\\Expected duration in a city\\Expected city on a day} & \makecell[tl]{Start time and place\\Travel time of two places\\Time and place of a person\\Min. duration with a person} & \makecell[tl]{Various provided clues\\about people and attributes}\\\bottomrule
    \end{tabular}%
    }
    \caption{Formal definition of variables $X_i$, domains $D_i$, and constraints $C_j$ from the 4 domains in \np and \zl.}
    \label{tab:naturalplan}
\end{table}

\section{Task and Data}
\label{sec:task_data}

To systematically compare \las and \laf, including various formalisms, we prioritize depth over breadth by focusing on real-world CSPs.
Such a problem includes the following elements described in natural language:
\begin{enumerate}[topsep=0pt,itemsep=-1ex,partopsep=1ex,parsep=1ex]
    \item A set of variables $\{X_i\}$, to be assigned values
    \item Domains of variables $\{D_i\}$, the possible values of the variables
    \item A set of constraints $\{C_j\}$, boolean rules that govern one or more variables
\end{enumerate}
The goal of planning is thus to find a value assignment of each variable $X_i$ based on possibilities $D_i$ that does not violate any constraints $C_j$.

We consider three domains, including calendar scheduling, trip planning, and meeting planning provided by the \np dataset \citep{zheng2024naturalplanbenchmarkingllms}, where only \las methods have been evaluated \citep{lee2025evolvingdeeperllmthinking,parmar2025plangenmultiagentframeworkgenerating} but not \laf.
We additionally consider a logic grid puzzle domain from the \zl dataset \citep{lin2025zebralogic}, which is less natural but considered in past work \citep{berman2024solvingzebrapuzzlesusing,kesseli2025logicpybridginggapllms} evaluating \laf methods.
The formulation and examples of all three tasks are shown in Table~\ref{tab:naturalplan}.
Unlike previous work that evaluated the predicted solution by matching the ground-truth solution, we manually and formally annotate all constraints $\{C_j\}$.
We therefore count a solution as correct if it formally satisfies all the constraints.
This design choice enables us to perform fine-grained analysis on problem complexity and model errors.
Bound by the cost of annotation, we randomly sample 100 out of 1,000 examples for each domain.
The tasks and prompts are exemplified in Appendix~\ref{sec:examples}.

\section{Experimental Setup}

We prompt LLMs in a one-shot manner to generate three different types of output.

\textbf{Solution.} This is the \las pipeline where the LLM is given an input and directly outputs the answer, optionally after generating a reasoning chain.
Noting that most of our LLMs are LRMs, we do not explicitly prompt any model to ``think step by step.''

\textbf{Python code.} This is an \laf pipeline where the LLM is given an input and generates a Python program that will be executed to output the answer.
Conceptually, the model generates both the declarative component (i.e., the variables and constraints) and the search component (i.e., the algorithm to search for the correct plan).

\textbf{SMT code.} Alternatively, the LLM may generate a specific program to invoke a CSP solver, such as an SMT solver.
Following previous work, we prompt LLMs to generate code using the Z3 solver Python wrapper.
Here, the model primarily generates the declarative component as the search component is simply a call to the pre-defined solver function.
Examples of the two formalisms are juxtaposed in Appendix~\ref{sec:python-z3}.

In all cases, the provided one-shot only illustrates the input-output format, containing neither an example solution nor an example formalized code. 
To not discount the \laf pipelines, we follow \citet{pan-etal-2023-logic,10.5555/3666122.3668096} to optionally include a revision-by-error module.
If the solver, either implemented by the model in the case of Python or provided by the Z3 library in the case of SMT, returns an error message or cannot find any plan, this signal is returned to the LLM to re-generate the program up to 5 times (see prompt examples in \autoref{sec:examples}).
We do not consider specific-purpose formalization pipelines such as \citep{hao2025planning} which defeats our purpose of studying general applications. We do not consider more involved \las pipelines as our goal is to use it as a lower bound to benchmark \laf.

\begin{figure}[htbp]
    \centering
    \small
    
    % calendar
    \begin{tikzpicture}
        \begin{axis}[
            ybar,
            bar width=3.5pt,
            width=\columnwidth,
            height=4.2cm,
            symbolic x coords={Qwen2.5*, DeepSeek-V3*, Qwen3, DeepSeek-R1, o3-mini, gpt-5},
            xtick=data,
            xticklabel style={rotate=30, anchor=east},
            enlarge x limits=0.15,
            ymin=0,
            ymax=100,
            title={Calendar Scheduling},
            legend image code/.code={
              \draw[fill=#1, draw=none] (0cm,-0.1cm) rectangle (0.3cm,0.1cm);
            }
        ]
        \addplot+[ybar, fill=red!50, draw=none] coordinates {(Qwen2.5*, 32) (DeepSeek-V3*, 99) (Qwen3, 97) (DeepSeek-R1, 100) (o3-mini, 99) (gpt-5, 100)};
        \addplot+[ybar, fill=blue!60, draw=none] coordinates {(Qwen2.5*, 56) (DeepSeek-V3*, 67) (Qwen3, 94) (DeepSeek-R1, 94) (o3-mini, 94) (gpt-5, 97)};
        \addplot+[ybar, fill=blue!70, draw=none] coordinates {(Qwen2.5*, 76) (DeepSeek-V3*, 91) (Qwen3, 96) (DeepSeek-R1, 95) (o3-mini, 94) (gpt-5, 97)};
        \addplot+[ybar, fill=cyan!60, draw=none] coordinates {(Qwen2.5*, 37) (DeepSeek-V3*, 85) (Qwen3, 90) (DeepSeek-R1, 91) (o3-mini, 84) (gpt-5, 77)};
        \addplot+[ybar, fill=cyan!90, draw=none] coordinates {(Qwen2.5*, 51) (DeepSeek-V3*, 96) (Qwen3, 96) (DeepSeek-R1, 99) (o3-mini, 100) (gpt-5, 100)};
        \end{axis}
    \end{tikzpicture}

    \vspace{0.2cm}

    % trip
    \begin{tikzpicture}
        \begin{axis}[
            ybar,
            bar width=3.5pt,
            width=\columnwidth,
            height=4.5cm,
            symbolic x coords={Qwen2.5*, DeepSeek-V3*, Qwen3, DeepSeek-R1, o3-mini, gpt-5},
            xtick=data,
            xticklabel style={rotate=30, anchor=east},
            enlarge x limits=0.15,
            ymin=0,
            ymax=100,
            title={Trip Planning},
            legend image code/.code={
              \draw[fill=#1, draw=none] (0cm,-0.1cm) rectangle (0.3cm,0.1cm);
            },
            legend style={
                at={(0.35,0.95)}, 
                anchor=north, 
                legend columns=5, 
                font=\scriptsize,
                /tikz/every even column/.append style={column sep=0.2cm}
            }
        ]
        \addplot+[ybar, fill=red!50, draw=none] coordinates {(Qwen2.5*, 12) (DeepSeek-V3*, 36) (Qwen3, 54) (DeepSeek-R1, 74) (o3-mini, 81) (gpt-5, 93)};
        \addplot+[ybar, fill=blue!60, draw=none] coordinates {(Qwen2.5*, 5) (DeepSeek-V3*, 6) (Qwen3, 43) (DeepSeek-R1, 64) (o3-mini, 72) (gpt-5, 83)};
        \addplot+[ybar, fill=blue!70, draw=none] coordinates {(Qwen2.5*, 5) (DeepSeek-V3*, 6) (Qwen3, 44) (DeepSeek-R1, 64) (o3-mini, 72) (gpt-5, 85)};
        \addplot+[ybar, fill=cyan!60, draw=none] coordinates {(Qwen2.5*, 0) (DeepSeek-V3*, 0) (Qwen3, 27) (DeepSeek-R1, 23) (o3-mini, 60) (gpt-5, 38)};
        \addplot+[ybar, fill=cyan!90, draw=none] coordinates {(Qwen2.5*, 0) (DeepSeek-V3*, 2) (Qwen3, 33) (DeepSeek-R1, 39) (o3-mini, 66) (gpt-5, 42)};
        \legend{Solution, Python, Python+rev, SMT, SMT+rev}
        \end{axis}
    \end{tikzpicture}

    \vspace{0.2cm}

    % meeting
    \begin{tikzpicture}
        \begin{axis}[
            ybar,
            bar width=3.5pt,
            width=\columnwidth,
            height=4.2cm,
            symbolic x coords={Qwen2.5*, DeepSeek-V3*, Qwen3, DeepSeek-R1, o3-mini, gpt-5},
            xtick=data,
            xticklabel style={rotate=30, anchor=east},
            enlarge x limits=0.15,
            ymin=0,
            ymax=100,
            title={Meeting Planning},
        ]
        \addplot+[ybar, fill=red!50, draw=none] coordinates {(Qwen2.5*, 18) (DeepSeek-V3*, 66) (Qwen3, 68) (DeepSeek-R1, 96) (o3-mini, 89) (gpt-5, 93)};
        \addplot+[ybar, fill=blue!60, draw=none] coordinates {(Qwen2.5*, 23) (DeepSeek-V3*, 82) (Qwen3, 56) (DeepSeek-R1, 97) (o3-mini, 82) (gpt-5, 95)};
        \addplot+[ybar, fill=blue!70, draw=none] coordinates {(Qwen2.5*, 26) (DeepSeek-V3*, 86) (Qwen3, 56) (DeepSeek-R1, 97) (o3-mini, 96) (gpt-5, 99)};
        \addplot+[ybar, fill=cyan!60, draw=none] coordinates {(Qwen2.5*, 15) (DeepSeek-V3*, 41) (Qwen3, 29) (DeepSeek-R1, 63) (o3-mini, 43) (gpt-5, 42)};
        \addplot+[ybar, fill=cyan!90, draw=none] coordinates {(Qwen2.5*, 23) (DeepSeek-V3*, 53) (Qwen3, 29) (DeepSeek-R1, 85) (o3-mini, 99) (gpt-5, 98)};
        \end{axis}
    \end{tikzpicture}

    \vspace{0.2cm}

    % zebra
    \begin{tikzpicture}
        \begin{axis}[
            ybar,
            bar width=3.5pt,
            width=\columnwidth,
            height=4.2cm,
            symbolic x coords={Qwen2.5*, DeepSeek-V3*, Qwen3, DeepSeek-R1, o3-mini, gpt-5},
            xtick=data,
            xticklabel style={rotate=30, anchor=east},
            enlarge x limits=0.15,
            ymin=0,
            ymax=100,
            title={Zebra Logic}
        ]
        \addplot+[ybar, fill=red!50, draw=none] coordinates {(Qwen2.5*, 21) (DeepSeek-V3*, 89) (Qwen3, 92) (DeepSeek-R1, 97) (o3-mini, 94) (gpt-5, 100)};
        \addplot+[ybar, fill=blue!60, draw=none] coordinates {(Qwen2.5*, 15) (DeepSeek-V3*, 81) (Qwen3, 87) (DeepSeek-R1, 86) (o3-mini, 92) (gpt-5, 93)};
        \addplot+[ybar, fill=blue!70, draw=none] coordinates {(Qwen2.5*, 32) (DeepSeek-V3*, 82) (Qwen3, 87) (DeepSeek-R1, 89) (o3-mini, 92) (gpt-5, 95)};
        \addplot+[ybar, fill=cyan!60, draw=none] coordinates {(Qwen2.5*, 15) (DeepSeek-V3*, 45) (Qwen3, 73) (DeepSeek-R1, 54) (o3-mini, 87) (gpt-5, 56)};
        \addplot+[ybar, fill=cyan!90, draw=none] coordinates {(Qwen2.5*, 32) (DeepSeek-V3*, 85) (Qwen3, 81) (DeepSeek-R1, 81) (o3-mini, 94) (gpt-5, 90)};
        \end{axis}
    \end{tikzpicture}
    
    \caption{The percentage of correct plans (defined by formally passing all annotated constraints) by \las and \laf generating both Python and SMT code of various LLMs on all 4 CSP domains. In the settings with revision, only solver errors, including the inability to find a plan, induce revisions. LLMs that are not LRMs are marked by an asterisk (*).}
    \label{fig:accuracy}
\end{figure}

\section{Results}

We present our findings guided by a series of research questions.

\subsection{RQ1: Are LLMs better formalizers than solvers?} 
\label{sec:rq1}
As discussed in Section~\ref{sec:relwork-laf}, much existing work has shown superior performance of \laf on various logical reasoning tasks. However, we observe that \textbf{LLM-as-formalizer underperforms LLM-as-solver} in 12 out of 16 LRM-dataset combinations and 4 out of 8 non-reasoning-LLM-dataset combinations (Figure~\ref{fig:accuracy}). In datasets with more saturated performance such as Calendar Scheduling, Meeting Planning, and Zebra Logic, models' formalization performance is decent but does not exceed the straightforward reasoning towards a solution in most cases. On the more challenging Trip Planning dataset, \laf consistently underperforms \las. The performance of the two methods roughly correlates, as an LLM that is a better solver is often also a better formalizer, even when comparing an LRM like \texttt{DeepSeek-R1} and a non-reasoning LLM like \texttt{DeepSeek-V3}. In contrast, the weakest model \texttt{Qwen2.5} as formalizer outperforms that as a solver. Notably, revision during formalization induces consistent performance gain in many cases, at the cost of more calls to the LLM. Comparing formalization targets, \textbf{general Python outperforms code to an SMT solver} in 15 out of 24 model-dataset combinations, even though SMT is a tool arguably more suitable for the CSP tasks providing much more syntactic convenience, while general Python is a higher-resource language during pre-training. 

\subsection{RQ2: Are formalizers more robust to complexity than solvers?} 
\label{sec:rq2}
\begin{figure}[t!]
    \centering
    \small
    \resizebox{\columnwidth}{!}{%
    \begin{tikzpicture}
      \begin{groupplot}[
        group style={
          group size=4 by 4,
          horizontal sep=0.1cm,
          vertical sep=0.1cm
        },
        width=5cm,
        height=4.2cm,
        xtick={1,2,3,4,5},
        ymin=0, ymax=100,
        ytick={0,25,50,75},
        grid=both,
        grid style={line width=.1pt, draw=gray!20},
        major grid style={line width=.2pt,draw=gray!50},
        legend pos=north east,
        legend style={
          font=\scriptsize,
          nodes={scale=1,transform shape},
        },
        legend image post style={scale=0.7},
        % remove axis labels
        xlabel={}, ylabel={},
      ]

      % Q2.5 calendar
      \nextgroupplot[title={Calendar Scheduling},xticklabels={},ylabel={Qwen2.5*}]
        \addplot+[mark=o, thick, draw=red!50] coordinates {
          (1,35) (2,30) (3,40) (4,25) (5,30)
        };
        \addplot+[mark=square, thick, draw=blue!70] coordinates {
          (1,80) (2,75) (3,65) (4,75) (5,85)
        };
        \addplot+[mark=triangle, thick, draw=cyan!90] coordinates {
          (1,60) (2,45) (3,45) (4,55) (5,50)
        };

      % Q2.5 trip
      \nextgroupplot[title={Trip Planning},xticklabels={},yticklabels={}]
        \addplot+[mark=o, thick, draw=red!50] coordinates {
          (1,40) (2,10) (3,10) (4,00) (5,00)
        };
        \addplot+[mark=square, thick, draw=blue!70] coordinates {
          (1,25) (2,00) (3,00) (4,00) (5,00)
        };
        \addplot+[mark=triangle, thick, draw=cyan!90] coordinates {
          (1,00) (2,00) (3,00) (4,00) (5,00)
        };

      % Q2.5 meeting
      \nextgroupplot[title={Meeting Planning},xticklabels={},yticklabels={}]
        \addplot+[mark=o, thick, draw=red!50] coordinates {
          (1,55) (2,15) (3,15) (4,00) (5,05)
        };
        \addplot+[mark=square, thick, draw=blue!70] coordinates {
          (1,55) (2,50) (3,20) (4,05) (5,00)
        };
        \addplot+[mark=triangle, thick, draw=cyan!90] coordinates {
          (1,85) (2,15) (3,05) (4,05) (5,05)
        };

      % Q2.5 zebra
      \nextgroupplot[title={Zebra Logic},xticklabels={},yticklabels={}]
        \addplot+[mark=o, thick, draw=red!50] coordinates {
          (1,64) (2,00) (3,00) (4,00)
        };
        \addplot+[mark=square, thick, draw=blue!70] coordinates {
          (1,52) (2,38) (3,24) (4,00)
        };
        \addplot+[mark=triangle, thick, draw=cyan!90] coordinates {
          (1,39) (2,42) (3,28) (4,11)
        };
        \legend{Plan, Python+rev, SMT+rev}

      % DSV3 calendar
      \nextgroupplot[xticklabels={},ylabel={DeepSeek-V3*}]
        \addplot+[mark=o, thick, draw=red!50] coordinates {
          (1,100) (2,100) (3,100) (4,100) (5,95)
        };
        \addplot+[mark=square, thick, draw=blue!70] coordinates {
          (1,95) (2,95) (3,90) (4,90) (5,85)
        };
        \addplot+[mark=triangle, thick, draw=cyan!90] coordinates {
          (1,100) (2,95) (3,95) (4,95) (5,95)
        };

      % DSV3 trip
      \nextgroupplot[yticklabels={},xticklabels={}]
        \addplot+[mark=o, thick, draw=red!50] coordinates {
          (1,75) (2,60) (3,20) (4,20) (5,05)
        };
        \addplot+[mark=square, thick, draw=blue!70] coordinates {
          (1,25) (2,00) (3,05) (4,00) (5,00)
        };
        \addplot+[mark=triangle, thick, draw=cyan!90] coordinates {
          (1,10) (2,00) (3,00) (4,00) (5,00)
        };

      % DSV3 meeting
      \nextgroupplot[yticklabels={},xticklabels={}]
        \addplot+[mark=o, thick, draw=red!50] coordinates {
          (1,95) (2,80) (3,90) (4,50) (5,15)
        };
        \addplot+[mark=square, thick, draw=blue!70] coordinates {
          (1,100) (2,80) (3,80) (4,90) (5,80)
        };
        \addplot+[mark=triangle, thick, draw=cyan!90] coordinates {
          (1,90) (2,55) (3,45) (4,25) (5,50)
        };

      % DSV3 zebra
      \nextgroupplot[yticklabels={},xticklabels={}]
        \addplot+[mark=o, thick, draw=red!50] coordinates {
          (1,100) (2,100) (3,80) (4,67)
        };
        \addplot+[mark=square, thick, draw=blue!70] coordinates {
          (1,97) (2,92) (3,100) (4,17)
        };
        \addplot+[mark=triangle, thick, draw=cyan!90] coordinates {
          (1,91) (2,92) (3,84) (4,67)
        };

      % Q3 calendar
      \nextgroupplot[xticklabels={},ylabel={Qwen3}]
        \addplot+[mark=o, thick, draw=red!50] coordinates {
          (1,100) (2,100) (3,95) (4,90) (5,100)
        };
        \addplot+[mark=square, thick, draw=blue!70] coordinates {
          (1,95) (2,100) (3,100) (4,95) (5,90)
        };
        \addplot+[mark=triangle, thick, draw=cyan!90] coordinates {
          (1,95) (2,100) (3,95) (4,95) (5,95)
        };

      % Q3 trip
      \nextgroupplot[yticklabels={},xticklabels={}]
        \addplot+[mark=o, thick, draw=red!50] coordinates {
          (1,75) (2,75) (3,70) (4,40) (5,10)
        };
        \addplot+[mark=square, thick, draw=blue!70] coordinates {
          (1,85) (2,75) (3,40) (4,20) (5,00)
        };
        \addplot+[mark=triangle, thick, draw=cyan!90] coordinates {
          (1,80) (2,55) (3,10) (4,20) (5,00)
        };

      % Q3 meeting
      \nextgroupplot[yticklabels={},xticklabels={}]
        \addplot+[mark=o, thick, draw=red!50] coordinates {
          (1,90) (2,90) (3,80) (4,40) (5,40)
        };
        \addplot+[mark=square, thick, draw=blue!70] coordinates {
          (1,75) (2,70) (3,45) (4,55) (5,35)
        };
        \addplot+[mark=triangle, thick, draw=cyan!90] coordinates {
          (1,80) (2,55) (3,10) (4,00) (5,00)
        };

      % Q3 zebra
      \nextgroupplot[yticklabels={},xticklabels={}]
        \addplot+[mark=o, thick, draw=red!50] coordinates {
          (1,100) (2,100) (3,96) (4,61)
        };
        \addplot+[mark=square, thick, draw=blue!70] coordinates {
          (1,97) (2,100) (3,92) (4,44)
        };
        \addplot+[mark=triangle, thick, draw=cyan!90] coordinates {
          (1,91) (2,83) (3,80) (4,61)
        };

      % DSR1 calendar
      \nextgroupplot[ylabel={DeepSeek-R1}]
        \addplot+[mark=o, thick, draw=red!50] coordinates {
          (1,100) (2,100) (3,100) (4,100) (5,95)
        };
        \addplot+[mark=square, thick, draw=blue!70] coordinates {
          (1,95) (2,95) (3,90) (4,90) (5,85)
        };
        \addplot+[mark=triangle, thick, draw=cyan!90] coordinates {
          (1,100) (2,95) (3,95) (4,95) (5,95)
        };

      % DSR1 trip
      \nextgroupplot[yticklabels={}]
        \addplot+[mark=o, thick, draw=red!50] coordinates {
          (1,75) (2,60) (3,20) (4,20) (5,05)
        };
        \addplot+[mark=square, thick, draw=blue!70] coordinates {
          (1,25) (2,00) (3,05) (4,00) (5,00)
        };
        \addplot+[mark=triangle, thick, draw=cyan!90] coordinates {
          (1,70) (2,40) (3,45) (4,15) (5,25)
        };

      % DSR1 meeting
      \nextgroupplot[yticklabels={}]
        \addplot+[mark=o, thick, draw=red!50] coordinates {
          (1,95) (2,80) (3,90) (4,50) (5,15)
        };
        \addplot+[mark=square, thick, draw=blue!70] coordinates {
          (1,95) (2,85) (3,95) (4,100) (5,95)
        };
        \addplot+[mark=triangle, thick, draw=cyan!90] coordinates {
          (1,95) (2,95) (3,70) (4,85) (5,80)
        };

      % DSR1 zebra
      \nextgroupplot[yticklabels={}]
        \addplot+[mark=o, thick, draw=red!50] coordinates {
          (1,100) (2,100) (3,80) (4,67)
        };
        \addplot+[mark=square, thick, draw=blue!70] coordinates {
          (1,97) (2,92) (3,100) (4,17)
        };
        \addplot+[mark=triangle, thick, draw=cyan!90] coordinates {
          (1,91) (2,92) (3,84) (4,67)
        };

    \end{groupplot}
  \end{tikzpicture}%
  }
  \caption{The change in the percentage of correct plans by \las and \laf over buckets of examples stratified by complexity measured by the number of constraints. Data in the \np domains is roughly equally partitioned into 5 percentiles, while that in \zl is partitioned based on the 4 strata provided by the dataset.}
  \label{fig:complexity}
\end{figure}
Even though \laf underperforms \las, the former is expected to be more robust to increasing problem complexity than \las because the two paradigms scale differently in their output structure. End-to-end solving requires the LLM to navigate the search space which exponentially explodes with the increased variables and constraints. Even though LRM was introdeuced to address this issue, existing work \citep{valmeekam2024llmscantplanlrms,shojaee2025illusionthinkingunderstandingstrengths} still pointed out their failure to scale as a solver when the search space of a problem becomes large. In contrast, formalization in theory decouples problem description from solution via declarative programming, leaving the heavy-lifting of search to the solver or the algorithmic component of the output code. Its workload is therefore linear to the input length, leading the community to expect substantially better robustness to complexity compared to \las. 

We validate the lack of robustness of LRMs as solvers (Figure~\ref{fig:complexity}) which quickly fail as the problems become complex in all domains other than the easiest Calendar Scheduling which we exclude from the discussion below. However, we note that \textbf{neither does \laf scale well with complexity}. For the better performing Python formalization, even with revision, the performance on the problems with top 20\% most constraints is less than half of the top 20\% least constraints in 10 out of 12 model-dataset combinations we study. Clearly, even with state-of-the-art LLMs and revision attempts, \laf does not show the promised robustness, pointing towards open space for improvement. 

\begin{figure}[htbp]
    \centering
    \small
    \begin{adjustbox}{width=\columnwidth}
    \begin{tikzpicture}

    % Legend
    \begin{scope}[shift={(-5.3, -2.5)}]
        \draw[fill=red!20] (-2, 0) rectangle ++(0.3, 0.3);
        \node[right] at (-1.6, 0.16) {Error};
        
        \draw[fill=brown!30] (-0.3, 0) rectangle ++(0.3, 0.3);
        \node[right] at (0.1, 0.16) {No plan};
        
        \draw[fill=blue!40] (1.6, 0) rectangle ++(0.3, 0.3);
        \node[right] at (1.9, 0.16) {Wrong plan};
        
        \draw[fill=green!40] (4, 0) rectangle ++(0.3, 0.3);
        \node[right] at (4.4, 0.16) {Correct};
    \end{scope}

    % Q3 Python Trip
    \begin{scope}[xshift=-10cm, yshift=1.5cm]
        \node[align=center] at (0, 1.5) {Qwen3, Py, Trip};
        \node[align=center, rotate=90] at (-1.5cm, 0cm) {No revision};
        \draw[fill=red!20] (0,0) circle (1cm);
        \draw[fill=brown!20] (0,0) circle (0.90cm);
        \draw[fill=blue!30] (0,0) circle (0.74cm); 
        \draw[fill=green!40] (0,0) circle (0.43cm);
    \end{scope}

    % Q3 SMT Trip
    \begin{scope}[xshift=-7.5cm, yshift=1.5cm]
        \node[align=center] at (0, 1.5) {Qwen3, SMT, Trip};
        \draw[fill=red!20] (0,0) circle (1cm);
        \draw[fill=brown!20] (0,0) circle (1cm);
        \draw[fill=blue!30] (0,0) circle (0.45cm);
        \draw[fill=green!40] (0,0) circle (0.27cm); 
    \end{scope}

    % DSR1 Python Trip
    \begin{scope}[xshift=-5cm, yshift=1.5cm]
        \node[align=center] at (0, 1.5) {R1, Py, Trip};
        \draw[fill=red!20] (0,0) circle (1cm);
        \draw[fill=brown!20] (0,0) circle (0.98cm);
        \draw[fill=blue!30] (0,0) circle (0.93cm);
        \draw[fill=green!40] (0,0) circle (0.64cm); 
    \end{scope}

    % R1 SMT Trip
    \begin{scope}[xshift=-2.5cm, yshift=1.5cm]
        \node[align=center] at (0, 1.4) {R1, SMT, Trip};
        \draw[fill=red!20] (0,0) circle (1cm);
        \draw[fill=brown!20] (0,0) circle (0.99cm);
        \draw[fill=blue!30] (0,0) circle (0.45cm);
        \draw[fill=green!40] (0,0) circle (0.23cm);
    \end{scope}

    % R1 Python Meeting
    \begin{scope}[yshift=1.5cm]
        \node[align=center] at (0, 1.4) {R1, Py, Meeting};
        \draw[fill=red!20] (0,0) circle (1cm);
        \draw[fill=brown!20] (0,0) circle (1cm);
        \draw[fill=blue!30] (0,0) circle (1cm);
        \draw[fill=green!40] (0,0) circle (.98cm);
    \end{scope}

    % R1 SMT Meeting
    \begin{scope}[xshift=2.5cm, yshift=1.5cm]
        \node[align=center] at (0, 1.4) {R1, SMT, Meeting};
        \draw[fill=red!20] (0,0) circle (1cm);
        \draw[fill=brown!20] (0,0) circle (.9cm);
        \draw[fill=blue!30] (0,0) circle (.75cm);
        \draw[fill=green!40] (0,0) circle (.707cm);
    \end{scope}

    % Q3 Python+rev Trip
    \begin{scope}[xshift=-10cm, yshift=-0.9cm]
        \node[align=center] at (0, 1.5) {};
        \node[align=center, rotate=90] at (-1.5cm, 0cm) {Revision};
        \draw[fill=red!20] (0,0) circle (1cm);
        \draw[fill=brown!20] (0,0) circle (0.99cm);
        \draw[fill=blue!30] (0,0) circle (0.93cm); 
        \draw[fill=green!40] (0,0) circle (0.44cm); 
    \end{scope}

    % Q3 SMT+rev Trip
    \begin{scope}[xshift=-7.5cm, yshift=-0.9cm]
        \node[align=center] at (0, 1.5) {};
        \draw[fill=red!20] (0,0) circle (1cm);
        \draw[fill=brown!20] (0,0) circle (1cm);
        \draw[fill=blue!30] (0,0) circle (0.78cm); 
        \draw[fill=green!40] (0,0) circle (0.33cm); 
    \end{scope}

    % R1 Python+rev Trip
    \begin{scope}[xshift=-5cm, yshift=-0.9cm]
        \node[align=center] at (0, 1.5) {};
        \draw[fill=red!20] (0,0) circle (1cm);
        \draw[fill=brown!20] (0,0) circle (1cm);
        \draw[fill=blue!30] (0,0) circle (0.99cm);
        \draw[fill=green!40] (0,0) circle (0.64cm); 
    \end{scope}

    % R1 SMT+rev Trip  NEEDS RERUN -> UPDATE: RE-RUN DONE
    \begin{scope}[xshift=-2.5cm, yshift=-0.9cm]
        \node[align=center] at (0, 1.4) {};
        \draw[fill=red!20] (0,0) circle (1cm);
        \draw[fill=brown!20] (0,0) circle (1cm);
        \draw[fill=blue!30] (0,0) circle (0.92cm);
        \draw[fill=green!40] (0,0) circle (0.39cm);
    \end{scope}

    % R1 Python+rev Meeting
    \begin{scope}[yshift=-0.9cm]
        \node[align=center] at (0, 1.4) {};
        \draw[fill=red!20] (0,0) circle (1cm);
        \draw[fill=brown!20] (0,0) circle (1cm);
        \draw[fill=blue!30] (0,0) circle (1cm);
        \draw[fill=green!40] (0,0) circle (.984cm);
    \end{scope}

    % DSR1 SMT+rev Meeting
    \begin{scope}[xshift=2.5cm, yshift=-0.9cm]
        \node[align=center] at (0, 1.4) {};
        \draw[fill=red!20] (0,0) circle (1cm);
        \draw[fill=brown!20] (0,0) circle (1cm);
        \draw[fill=blue!30] (0,0) circle (0.95cm);
        \draw[fill=green!40] (0,0) circle (0.9cm);
    \end{scope}

    \end{tikzpicture}
    \end{adjustbox}
    \caption{Error analysis of two LRMs \texttt{Qwen3} and \texttt{DeepSeek-R1} as both Python and SMT formalizers. Revision is only allowed in case of errors (red ring) and not being able to find a plan (brown ring), and thus has no bearing on existing wrong plans (blue ring).}
    \label{fig:code_error}
\end{figure}

\subsection{RQ3: Why does \laf underperform and fail to scale?}
\label{sec:rq3}
To understand why LLMs are struggling to generate syntactically and semantically formal programs, we consider 4 outcomes:
\begin{itemize}
    \item \textbf{Error}: the program is unable to be executed due to an error such as TypeError, RuntimeError, or an error specific to the Z3 library.
    \item \textbf{No plan}: the solver outputs a custom message that no plan can be found.
    \item  \textbf{Wrong plan}: the solver outputs a plan, but the plan does not pass all the constraint checks for \np or does not exactly match the ground truth for \zl.
    \item \textbf{Correct plan}: the plan is correct.
\end{itemize}

Figure~\ref{fig:code_error} presents the error analysis of \laf conducted of on Trip Planning and Meeting Planning. We focus on these two open-source LRMs with distinct architectures (Mixture-of-Experts and dense) and on these two tasks, since they exhibit the strongest contrast in outcomes, with \las consistently outperforming \laf on Trip Planning and vice versa in Meeting Planning.

Syntax errors are rare but still existent, while the majority of them are eliminated given revision. Common syntax errors are diverse and include import errors, value error, and run-time errors, but are primarily dominated by erroneous use of APIs in the Z3 library for SMT generation. Similarly, the inability to find a plan, which is a kind of semantic errors, can be largely addressed with revision. With its effect on both kinds of errors, revision does increase overall correctness of the solution. The persistent and challenging error is a wrong plan, which is associated with capabilities of translating constraints and conducting correct algorithmic reasoning. In real-life applications, validating the correctness of plans assumes much more resources and risks than validating the existence of a plan. 

To further study the root cause of semantic errors, including both cases of ``no plan'' and ``wrong plan'', we manually annotate 5 examples per domain, per Python or SMT, per \texttt{Qwen3} or \texttt{gpt-5} as a formalizer, 80 examples in total. We break down the semantic errors into 3 finer-grained categories:
\begin{itemize}
    \item \textbf{Missing constraint}: the program misses the definition of a constraint otherwise present in the problem description.
    \item \textbf{Wrong constraint}: the program wrongly defines a constraint as presented in the problem description.
    \item  \textbf{Wrong solver}: the program wrongly defines or calls the solver.
\end{itemize}

We count that across domains and formalisms, \textbf{the majority of the semantic errors are due to wrongly defining constraints} (95\% of \texttt{Qwen3}). Examples include treating $2:16$ as an end time while it should be the start time for Calendar Scheduling, defining meeting time to be $9:16$ while it should be $9:18$ for Meeting plan, and so on. Such trivial errors of information extraction still plague the state-of-the-art LRMs and cannot be caught by the error message provided by the solver, posing a great safety concern in real-life applications. This calls for the need for a specific translation module beyond solely relying on LLMs to perform the translation from the problem description to the formal program. The other two categories, missing constraint and wrong solver, are much less common, though still existent. 

\begin{figure}[htbp]
  \centering
  \small
  \resizebox{\columnwidth}{!}{%
  \begin{tikzpicture}
    \begin{groupplot}[
      group style={
        group size=4 by 1,       % 4 plots in one row
        horizontal sep=0.2cm,    % Tightened horizontal spacing
        vertical sep=1.5cm
      },
      width=4.5cm,               % Slightly narrower width per plot before scaling
      height=4cm,                % Adjusted height
      ybar=2pt,                  % bar chart style
      ymin=0,
      ymax=14,                   % Increased from 10 so the 13.032 value isn't clipped
      enlarge x limits=0.5,      % spacing at sides
      symbolic x coords={Qwen3,DeepSeek-R1},
      xtick=data,
      xticklabel style={rotate=25, anchor=east}, % Rotated slightly to prevent overlap
      every axis plot/.append style={/tikz/bar width=6pt},  % set bar width here
      legend image code/.code={
        \draw[fill=#1, draw=none] (0cm,-0.1cm) rectangle (0.3cm,0.1cm);
      },
      legend pos=north east,
      legend style={font=\tiny}
    ]

      \nextgroupplot[title={Calendar Scheduling}]
      \addplot+[fill=red!50,draw=none]  coordinates {(Qwen3,4.267) (DeepSeek-R1,2.989)};
      \addplot+[fill=blue!70,draw=none]   coordinates {(Qwen3,3.094) (DeepSeek-R1,1.327)};
      \addplot+[fill=cyan!90,draw=none]   coordinates {(Qwen3,3.427) (DeepSeek-R1,0.989)};
      \legend{Plan, Python, SMT}

      \nextgroupplot[title={Trip Planning}, yticklabels={}]
      \addplot+[fill=red!50,draw=none]  coordinates {(Qwen3,6.156) (DeepSeek-R1,13.032)};
      \addplot+[fill=blue!70,draw=none]   coordinates {(Qwen3,3.551) (DeepSeek-R1,0.961)};
      \addplot+[fill=cyan!90,draw=none]   coordinates {(Qwen3,3.793) (DeepSeek-R1,1.272)};

      \nextgroupplot[title={Meeting Planning}, yticklabels={}] % Added yticklabels={} here
      \addplot+[fill=red!50,draw=none]  coordinates {(Qwen3,5.541) (DeepSeek-R1,7.558)};
      \addplot+[fill=blue!70,draw=none]   coordinates {(Qwen3,3.818) (DeepSeek-R1,1.846)};
      \addplot+[fill=cyan!90,draw=none]   coordinates {(Qwen3,4.123) (DeepSeek-R1,1.985)};

      \nextgroupplot[title={Zebra Logic}, yticklabels={}]
      \addplot+[fill=red!50,draw=none]  coordinates {(Qwen3,5.077) (DeepSeek-R1,3.544)};
      \addplot+[fill=blue!70,draw=none]   coordinates {(Qwen3,8.504) (DeepSeek-R1,1.237)};
      \addplot+[fill=cyan!90,draw=none]   coordinates {(Qwen3,8.786) (DeepSeek-R1,1.376)};

    \end{groupplot}
  \end{tikzpicture}%
  }
  \caption{Number of reasoning tokens (in thousands) generated by LRMs as a solver (Plan) and as a formalizer (Python and SMT).}
  \label{fig:reasoning_tokens}
\end{figure}
\begin{figure*}[ht!]
  \centering

  \begin{subfigure}[t]{0.24\linewidth}
    \centering
    \begin{tikzpicture}
      \pie[
        radius=1.0,
        text={},
        sum=auto
      ]{26/Sound, 10/Unnecessary, 4/Spurious}
    \end{tikzpicture}
    \caption{Qwen3, Python}
  \end{subfigure}%
  \hfill
  \begin{subfigure}[t]{0.24\linewidth}
    \centering
    \begin{tikzpicture}
      \pie[
        radius=1.0,
        text={},
        sum=auto
      ]{24/Sound, 10/Unnecessary, 6/Spurious}
    \end{tikzpicture}
    \caption{Qwen3, SMT}
  \end{subfigure}%
  \hfill
  \begin{subfigure}[t]{0.24\linewidth}
    \centering
    \begin{tikzpicture}
      \pie[
        radius=1.0,
        text={},
        sum=auto
      ]{20/Correct, 7/Unnecessary, 13/Spurious}
    \end{tikzpicture}
    \caption{DeepSeek-R1, Python}
  \end{subfigure}%
  \hfill
  \begin{subfigure}[t]{0.24\linewidth}
    \centering
    \begin{tikzpicture}
      \pie[
        radius=1.0,
        text={},
        sum=auto
      ]{3/Correct, 36/Unnecessary, 1/Spurious}
    \end{tikzpicture}
    \caption{DeepSeek-R1, SMT}
  \end{subfigure}

  \caption{The distribution of different types of reasoning chains (\textcolor{blue}{code-related}, \textcolor{cyan}{solver-like}, and \textcolor{yellow}{spurious}) in all domains for both LRMs and both formalisms.}
  \label{fig:reasoning-chain-analyis}
\end{figure*}
\begin{figure}[htbp]
\centering
\begin{lstlisting}[style=mypython]
# Reasoning chain: ... Let me think about a possible sequence. 
# Maybe the user should visit Rome on the first day? Wait, but 
# that will violate the constraint of being in Dubrovnik for a 
# conference. Let me think again...
import json
def main():
    itinerary = [
        {"day_range": "Day 1-2", "place": "Dubrovnik"},
        {"day_range": "Day 3", "place": "Rome (travel from Dubrovnik)"},
        {"day_range": "Day 4", "place": "Riga (travel from Rome)"},
        # More days omitted
    ]
    print(json.dumps({"itinerary": itinerary}))
if __name__ == "__main__":
    main()
\end{lstlisting}
\caption{An example of solver-like and spurious reasoning resulting in a hard-coded program (abridged) of \texttt{DeepSeek-R1} as a Python formalizer on Trip Planning.}
\label{fig:python-itinerary}
\end{figure}

\subsection{RQ4: Are LRMs more efficient as a formalizer than as a planner?}
\label{sec:rq4}

It is known that LRMs' impressive performance as a planner comes at a cost of excessive and often inefficient reasoning chains \citep{sui2025stop}. On the other hand, \laf may be expected to generate fewer reasoning tokens because the task of formalizing is $O(n+d)$ with $n$ variables and a domain size of $d$, as opposed to the worst case $O(d^n)$ for an NP-complete problem like CSP. We show that indeed \textbf{LRMs often generate significantly less reasoning tokens as a formalizer than as a solver}. From Figure~\ref{fig:reasoning_tokens}, it can be seen that the number of tokens is consistently higher for \las compared to \laf except the Zebra-Logic domain with \texttt{Qwen3}. We also observe that the number of tokens in the case of Python formalization is less than SMT formalization. 

In addition to the efficiency of the reasoning chains during formalization, we also inspect their nature by breaking them down into the following granular categories: 
\begin{itemize}
    \item \textbf{Code-related reasoning} focuses on coding such as errors, edge cases, and fixes.
    \item \textbf{Solver-like reasoning} unnecessarily tries to solve the problem like \las, but still generates code that declares the constraints and calls a solver.
    \item \textbf{Spurious reasoning} hard-codes the answer without the constraints and the solver within the code.
\end{itemize}

We manually annotate the reasoning chains of \texttt{DeepSeek-R1} and \texttt{Qwen3} for 10 examples per domain per formalism, resulting in 160 labels in total. Figure~\ref{fig:reasoning-chain-analyis} reveals a concerning phenomenon where only less than 75\% of the reasoning chains of \texttt{Qwen3} and less than 50\% of those of \texttt{DeepSeek-R1} are primarily code-related. \textbf{\laf frequently ``reason like a solver''} to perform search, enumeration, and backtracking to attempt to solve the problem rather than formalizing it. We count that the average number of tokens generated for such solver-like reasoning more than 1.5 times the token consumption than code-related reasoning. A likely explanation of this behavioral mismatch is that LRMs are trained to reason towards a solution, not formalize, and such behavior manifests when the the model is prompted to formalize, giving rise to suboptimal performance.

Despite those shortcomings, one of the built-in advantages of \laf is its improved interpretability, verifiability, and faithfulness, as previously discussed. However, spurious reasoning, the most insidious category of reasoning chains, nullifies this advantage. Figure~\ref{fig:python-itinerary} shows an example of spurious reasoning, where the model not only generates a solver-like, unnecessary reasoning chain, but also \textbf{hard-codes the entire proposed solution in the output program} without performing any declaration of constraints or implementation of search. These undesired behavioral deviations from the formalization instruction indicates a tendency that steers LLMs towards solving the problem directly instead of formalization. Regardless of the correctness, a practitioner would have no way of interpreting and verifying such a hard-coded program. Unnecessary reasoning chains constitutes as much as 90\% of programs generated by \texttt{Deepseek-R1} in all domains, revealing a considerable concern in safety. 

%\section{Discussion}
%In response to the current line of effort in the community to develop \laf, our findings from a systematic evaluation on CSP domains provide a more positive and nuanced view of its practical effectiveness.
%While \laf and \las achieve comparable accuracy across domains, \laf demonstrates clear advantages in efficiency and scalability.
%Its output length grows mainly with the number of variables and constraints, while the solver component remains fixed, resulting in fewer generated tokens and more stable behavior as problem complexity increases.
%In addition, \laf naturally offers robustness, verifiability, interpretability, and feasibility through explicit formal representations.

\section{Conclusion}
We provide a comprehensive evaluation of LLMs’ ability to formalize real-life constraint satisfaction problems and outsourcing the heavy-lifting of search to external tools. In contrastive to existing work that advocates for the efficacy of \laf, we show that it consistently underperforms \las on real-life constraint satisfaction problems, offering no more scalability with problem complexity. Our in-depth error analysis points to missing constraints and erroneous logic, internally caused by excessive, solver-like reasoning tokens that sometime lead to hard-coded solution. Our work indicates urgency in addressing these issues for \laf that otherwise promises verifiability, interpretability, and application in high-stakes domains. 

%\section*{Impact Statement}
%This work serves to evaluate large language models as formalizers for real-world constraint satisfaction problems.
%It shows that they achieve on-par accuracy with end-to-end LLM solvers, while some tasks even show higher efficiency in token usage and shorter reasoning.
%Such properties may benefit planning and scheduling applications in the field of LLM research, and can enable improvements in explicit constraint verification tasks.
%However, we find that current models can misinterpret constraints or even hard-code solutions, creating a possible risk when deploying without any oversight.
%The broader societal impact, therefore, depends on the caution taken during usage and the continued research into verification and safe deployment strategies when evaluating LLM-based formalizers.

\bibliography{anthology-1, anthology-2, colm2026_conference}

\begin{thebibliography}{43}
\providecommand{\natexlab}[1]{#1}
\providecommand{\url}[1]{\texttt{#1}}
\expandafter\ifx\csname urlstyle\endcsname\relax
  \providecommand{\doi}[1]{doi: #1}\else
  \providecommand{\doi}{doi: \begingroup \urlstyle{rm}\Url}\fi

\bibitem[Beiser et~al.(2025)Beiser, Penz, and Musliu]{beiser2025intermediatelanguagesmatterformal}
Alexander Beiser, David Penz, and Nysret Musliu.
\newblock Intermediate languages matter: Formal choice drives neurosymbolic llm reasoning, 2025.
\newblock URL \url{https://arxiv.org/abs/2502.17216}.

\bibitem[Berman et~al.(2024)Berman, McKeown, and Ray]{berman2024solvingzebrapuzzlesusing}
Shmuel Berman, Kathleen McKeown, and Baishakhi Ray.
\newblock Solving zebra puzzles using constraint-guided multi-agent systems, 2024.
\newblock URL \url{https://arxiv.org/abs/2407.03956}.

\bibitem[Brown et~al.(2020)Brown, Mann, Ryder, Subbiah, Kaplan, Dhariwal, Neelakantan, Shyam, Sastry, Askell, Agarwal, Herbert-Voss, Krueger, Henighan, Child, Ramesh, Ziegler, Wu, Winter, Hesse, Chen, Sigler, Litwin, Gray, Chess, Clark, Berner, McCandlish, Radford, Sutskever, and Amodei]{10.5555/3495724.3495883}
Tom~B. Brown, Benjamin Mann, Nick Ryder, Melanie Subbiah, Jared Kaplan, Prafulla Dhariwal, Arvind Neelakantan, Pranav Shyam, Girish Sastry, Amanda Askell, Sandhini Agarwal, Ariel Herbert-Voss, Gretchen Krueger, Tom Henighan, Rewon Child, Aditya Ramesh, Daniel~M. Ziegler, Jeffrey Wu, Clemens Winter, Christopher Hesse, Mark Chen, Eric Sigler, Mateusz Litwin, Scott Gray, Benjamin Chess, Jack Clark, Christopher Berner, Sam McCandlish, Alec Radford, Ilya Sutskever, and Dario Amodei.
\newblock Language models are few-shot learners.
\newblock In \emph{Proceedings of the 34th International Conference on Neural Information Processing Systems}, NIPS '20, Red Hook, NY, USA, 2020. Curran Associates Inc.
\newblock ISBN 9781713829546.

\bibitem[Chen et~al.(2025)Chen, Jhamtani, Sharma, Fan, and Wang]{chen2025steering}
Yongchao Chen, Harsh Jhamtani, Srinagesh Sharma, Chuchu Fan, and Chi Wang.
\newblock Steering large language models between code execution and textual reasoning.
\newblock In \emph{The Thirteenth International Conference on Learning Representations}, 2025.
\newblock URL \url{https://openreview.net/forum?id=5X5Z7Ffrjb}.

\bibitem[Clark et~al.(2021)Clark, Tafjord, and Richardson]{10.5555/3491440.3491977}
Peter Clark, Oyvind Tafjord, and Kyle Richardson.
\newblock Transformers as soft reasoners over language.
\newblock In \emph{Proceedings of the Twenty-Ninth International Joint Conference on Artificial Intelligence}, IJCAI'20, 2021.
\newblock ISBN 9780999241165.

\bibitem[Feng et~al.(2023)Feng, Zhang, Gu, Ye, He, and Wang]{feng2023towards}
Guhao Feng, Bohang Zhang, Yuntian Gu, Haotian Ye, Di~He, and Liwei Wang.
\newblock Towards revealing the mystery behind chain of thought: A theoretical perspective.
\newblock In \emph{Thirty-seventh Conference on Neural Information Processing Systems}, 2023.
\newblock URL \url{https://openreview.net/forum?id=qHrADgAdYu}.

\bibitem[Gao et~al.(2023)Gao, Madaan, Zhou, Alon, Liu, Yang, Callan, and Neubig]{10.5555/3618408.3618843}
Luyu Gao, Aman Madaan, Shuyan Zhou, Uri Alon, Pengfei Liu, Yiming Yang, Jamie Callan, and Graham Neubig.
\newblock Pal: program-aided language models.
\newblock In \emph{Proceedings of the 40th International Conference on Machine Learning}, ICML'23. JMLR.org, 2023.

\bibitem[Guo et~al.(2025)Guo, Yang, Zhang, Song, Wang, Zhu, Xu, Zhang, Ma, Bi, Zhang, Yu, Wu, Wu, Gou, Shao, Li, Gao, Liu, Xue, Wang, Wu, Feng, Lu, Zhao, Deng, Ruan, Dai, Chen, Ji, Li, Lin, Dai, Luo, Hao, Chen, Li, Zhang, Xu, Ding, Gao, Qu, Li, Guo, Li, Chen, Yuan, Tu, Qiu, Li, Cai, Ni, Liang, Chen, Dong, Hu, You, Gao, Guan, Huang, Yu, Wang, Zhang, Zhao, Wang, Zhang, Xu, Xia, Zhang, Zhang, Tang, Zhou, Li, Wang, Li, Tian, Huang, Zhang, Wang, Chen, Du, Ge, Zhang, Pan, Wang, Chen, Jin, Chen, Lu, Zhou, Chen, Ye, Wang, Yu, Zhou, Pan, Li, Zhou, Wu, Yun, Pei, Sun, Wang, Zeng, Liu, Liang, Gao, Yu, Zhang, Xiao, An, Liu, Wang, Chen, Nie, Cheng, Liu, Xie, Liu, Yang, Li, Su, Lin, Li, Jin, Shen, Chen, Sun, Wang, Song, Zhou, Wang, Shan, Li, Wang, Wei, Zhang, Xu, Li, Zhao, Sun, Wang, Yu, Zhang, Shi, Xiong, He, Piao, Wang, Tan, Ma, Liu, Guo, Ou, Wang, Gong, Zou, He, Xiong, Luo, You, Liu, Zhou, Zhu, Huang, Li, Zheng, Zhu, Ma, Tang, Zha, Yan, Ren, Ren, Sha, Fu, Xu, Xie, Zhang, Hao, Ma, Yan, Wu, Gu, Zhu, Liu, Li, Xie, Song,
  Pan, Huang, Xu, Zhang, and Zhang]{guo2025deepseek}
Daya Guo, Dejian Yang, Haowei Zhang, Junxiao Song, Peiyi Wang, Qihao Zhu, Runxin Xu, Ruoyu Zhang, Shirong Ma, Xiao Bi, Xiaokang Zhang, Xingkai Yu, Yu~Wu, Z.~F. Wu, Zhibin Gou, Zhihong Shao, Zhuoshu Li, Ziyi Gao, Aixin Liu, Bing Xue, Bingxuan Wang, Bochao Wu, Bei Feng, Chengda Lu, Chenggang Zhao, Chengqi Deng, Chong Ruan, Damai Dai, Deli Chen, Dongjie Ji, Erhang Li, Fangyun Lin, Fucong Dai, Fuli Luo, Guangbo Hao, Guanting Chen, Guowei Li, H.~Zhang, Hanwei Xu, Honghui Ding, Huazuo Gao, Hui Qu, Hui Li, Jianzhong Guo, Jiashi Li, Jingchang Chen, Jingyang Yuan, Jinhao Tu, Junjie Qiu, Junlong Li, J.~L. Cai, Jiaqi Ni, Jian Liang, Jin Chen, Kai Dong, Kai Hu, Kaichao You, Kaige Gao, Kang Guan, Kexin Huang, Kuai Yu, Lean Wang, Lecong Zhang, Liang Zhao, Litong Wang, Liyue Zhang, Lei Xu, Leyi Xia, Mingchuan Zhang, Minghua Zhang, Minghui Tang, Mingxu Zhou, Meng Li, Miaojun Wang, Mingming Li, Ning Tian, Panpan Huang, Peng Zhang, Qiancheng Wang, Qinyu Chen, Qiushi Du, Ruiqi Ge, Ruisong Zhang, Ruizhe Pan, Runji Wang, R.~J.
  Chen, R.~L. Jin, Ruyi Chen, Shanghao Lu, Shangyan Zhou, Shanhuang Chen, Shengfeng Ye, Shiyu Wang, Shuiping Yu, Shunfeng Zhou, Shuting Pan, S.~S. Li, Shuang Zhou, Shaoqing Wu, Tao Yun, Tian Pei, Tianyu Sun, T.~Wang, Wangding Zeng, Wen Liu, Wenfeng Liang, Wenjun Gao, Wenqin Yu, Wentao Zhang, W.~L. Xiao, Wei An, Xiaodong Liu, Xiaohan Wang, Xiaokang Chen, Xiaotao Nie, Xin Cheng, Xin Liu, Xin Xie, Xingchao Liu, Xinyu Yang, Xinyuan Li, Xuecheng Su, Xuheng Lin, X.~Q. Li, Xiangyue Jin, Xiaojin Shen, Xiaosha Chen, Xiaowen Sun, Xiaoxiang Wang, Xinnan Song, Xinyi Zhou, Xianzu Wang, Xinxia Shan, Y.~K. Li, Y.~Q. Wang, Y.~X. Wei, Yang Zhang, Yanhong Xu, Yao Li, Yao Zhao, Yaofeng Sun, Yaohui Wang, Yi~Yu, Yichao Zhang, Yifan Shi, Yiliang Xiong, Ying He, Yishi Piao, Yisong Wang, Yixuan Tan, Yiyang Ma, Yiyuan Liu, Yongqiang Guo, Yuan Ou, Yuduan Wang, Yue Gong, Yuheng Zou, Yujia He, Yunfan Xiong, Yuxiang Luo, Yuxiang You, Yuxuan Liu, Yuyang Zhou, Y.~X. Zhu, Yanping Huang, Yaohui Li, Yi~Zheng, Yuchen Zhu, Yunxian Ma, Ying
  Tang, Yukun Zha, Yuting Yan, Z.~Z. Ren, Zehui Ren, Zhangli Sha, Zhe Fu, Zhean Xu, Zhenda Xie, Zhengyan Zhang, Zhewen Hao, Zhicheng Ma, Zhigang Yan, Zhiyu Wu, Zihui Gu, Zijia Zhu, Zijun Liu, Zilin Li, Ziwei Xie, Ziyang Song, Zizheng Pan, Zhen Huang, Zhipeng Xu, Zhongyu Zhang, and Zhen Zhang.
\newblock Deepseek-r1 incentivizes reasoning in llms through reinforcement learning.
\newblock \emph{Nature}, 645\penalty0 (8081):\penalty0 633--638, Sep 2025.
\newblock ISSN 1476-4687.
\newblock \doi{10.1038/s41586-025-09422-z}.
\newblock URL \url{https://doi.org/10.1038/s41586-025-09422-z}.

\bibitem[Han et~al.(2024)Han, Schoelkopf, Zhao, Qi, Riddell, Zhou, Coady, Peng, Qiao, Benson, Sun, Wardle-Solano, Szab{\'o}, Zubova, Burtell, Fan, Liu, Wong, Sailor, Ni, Nan, Kasai, Yu, Zhang, Fabbri, Kryscinski, Yavuz, Liu, Lin, Joty, Zhou, Xiong, Ying, Cohan, and Radev]{han-etal-2024-folio}
Simeng Han, Hailey Schoelkopf, Yilun Zhao, Zhenting Qi, Martin Riddell, Wenfei Zhou, James Coady, David Peng, Yujie Qiao, Luke Benson, Lucy Sun, Alexander Wardle-Solano, Hannah Szab{\'o}, Ekaterina Zubova, Matthew Burtell, Jonathan Fan, Yixin Liu, Brian Wong, Malcolm Sailor, Ansong Ni, Linyong Nan, Jungo Kasai, Tao Yu, Rui Zhang, Alexander Fabbri, Wojciech~Maciej Kryscinski, Semih Yavuz, Ye~Liu, Xi~Victoria Lin, Shafiq Joty, Yingbo Zhou, Caiming Xiong, Rex Ying, Arman Cohan, and Dragomir Radev.
\newblock {FOLIO}: Natural language reasoning with first-order logic.
\newblock In Yaser Al-Onaizan, Mohit Bansal, and Yun-Nung Chen (eds.), \emph{Proceedings of the 2024 Conference on Empirical Methods in Natural Language Processing}, pp.\  22017--22031, Miami, Florida, USA, November 2024. Association for Computational Linguistics.
\newblock \doi{10.18653/v1/2024.emnlp-main.1229}.
\newblock URL \url{https://aclanthology.org/2024.emnlp-main.1229/}.

\bibitem[Hao et~al.(2025)Hao, Zhang, and Fan]{hao2025planning}
Yilun Hao, Yang Zhang, and Chuchu Fan.
\newblock Planning anything with rigor: General-purpose zero-shot planning with {LLM}-based formalized programming.
\newblock In \emph{The Thirteenth International Conference on Learning Representations}, 2025.
\newblock URL \url{https://openreview.net/forum?id=0K1OaL6XuK}.

\bibitem[Huang \& Zhang(2025)Huang and Zhang]{huang-zhang-2025-limit}
Cassie Huang and Li~Zhang.
\newblock On the limit of language models as planning formalizers.
\newblock In Wanxiang Che, Joyce Nabende, Ekaterina Shutova, and Mohammad~Taher Pilehvar (eds.), \emph{Proceedings of the 63rd Annual Meeting of the Association for Computational Linguistics (Volume 1: Long Papers)}, pp.\  4880--4904, Vienna, Austria, July 2025. Association for Computational Linguistics.
\newblock ISBN 979-8-89176-251-0.
\newblock \doi{10.18653/v1/2025.acl-long.242}.
\newblock URL \url{https://aclanthology.org/2025.acl-long.242/}.

\bibitem[Jha et~al.(2023)Jha, Jha, Lincoln, Bastian, Velasquez, and Neema]{10207581}
Susmit Jha, Sumit~Kumar Jha, Patrick Lincoln, Nathaniel~D. Bastian, Alvaro Velasquez, and Sandeep Neema.
\newblock Dehallucinating large language models using formal methods guided iterative prompting.
\newblock In \emph{2023 IEEE International Conference on Assured Autonomy (ICAA)}, pp.\  149--152, 2023.
\newblock \doi{10.1109/ICAA58325.2023.00029}.

\bibitem[Jiang et~al.(2023)Jiang, Welleck, Zhou, Lacroix, Liu, Li, Jamnik, Lample, and Wu]{jiang2023draft}
Albert~Qiaochu Jiang, Sean Welleck, Jin~Peng Zhou, Timothee Lacroix, Jiacheng Liu, Wenda Li, Mateja Jamnik, Guillaume Lample, and Yuhuai Wu.
\newblock Draft, sketch, and prove: Guiding formal theorem provers with informal proofs.
\newblock In \emph{The Eleventh International Conference on Learning Representations}, 2023.
\newblock URL \url{https://openreview.net/forum?id=SMa9EAovKMC}.

\bibitem[Kagitha et~al.(2025)Kagitha, Zhu, and Zhang]{kagitha2025addressingchallengesplanninglanguage}
Prabhu~Prakash Kagitha, Andrew Zhu, and Li~Zhang.
\newblock Addressing the challenges of planning language generation, 2025.
\newblock URL \url{https://arxiv.org/abs/2505.14763}.

\bibitem[Kesseli et~al.(2025)Kesseli, O'Hearn, and Cabral]{kesseli2025logicpybridginggapllms}
Pascal Kesseli, Peter O'Hearn, and Ricardo~Silveira Cabral.
\newblock Logic.py: Bridging the gap between llms and constraint solvers, 2025.
\newblock URL \url{https://arxiv.org/abs/2502.15776}.

\bibitem[Kojima et~al.(2022)Kojima, Gu, Reid, Matsuo, and Iwasawa]{10.5555/3600270.3601883}
Takeshi Kojima, Shixiang~Shane Gu, Machel Reid, Yutaka Matsuo, and Yusuke Iwasawa.
\newblock Large language models are zero-shot reasoners.
\newblock In \emph{Proceedings of the 36th International Conference on Neural Information Processing Systems}, NIPS '22, Red Hook, NY, USA, 2022. Curran Associates Inc.
\newblock ISBN 9781713871088.

\bibitem[Lee et~al.(2025)Lee, Fischer, Wu, Marwood, Baluja, Schuurmans, and Chen]{lee2025evolvingdeeperllmthinking}
Kuang-Huei Lee, Ian Fischer, Yueh-Hua Wu, Dave Marwood, Shumeet Baluja, Dale Schuurmans, and Xinyun Chen.
\newblock Evolving deeper llm thinking, 2025.
\newblock URL \url{https://arxiv.org/abs/2501.09891}.

\bibitem[Lin et~al.(2025)Lin, Bras, Richardson, Sabharwal, Poovendran, Clark, and Choi]{lin2025zebralogic}
Bill~Yuchen Lin, Ronan~Le Bras, Kyle Richardson, Ashish Sabharwal, Radha Poovendran, Peter Clark, and Yejin Choi.
\newblock Zebralogic: On the scaling limits of {LLM}s for logical reasoning.
\newblock In \emph{Forty-second International Conference on Machine Learning}, 2025.
\newblock URL \url{https://openreview.net/forum?id=sTAJ9QyA6l}.

\bibitem[Liu et~al.(2023)Liu, Jiang, Zhang, Liu, Zhang, Biswas, and Stone]{liu2023llm+}
Bo~Liu, Yuqian Jiang, Xiaohan Zhang, Qiang Liu, Shiqi Zhang, Joydeep Biswas, and Peter Stone.
\newblock Llm+ p: Empowering large language models with optimal planning proficiency.
\newblock \emph{arXiv preprint arXiv:2304.11477}, 2023.

\bibitem[Lyu et~al.(2023)Lyu, Havaldar, Stein, Zhang, Rao, Wong, Apidianaki, and Callison-Burch]{lyu-etal-2023-faithful}
Qing Lyu, Shreya Havaldar, Adam Stein, Li~Zhang, Delip Rao, Eric Wong, Marianna Apidianaki, and Chris Callison-Burch.
\newblock Faithful chain-of-thought reasoning.
\newblock In Jong~C. Park, Yuki Arase, Baotian Hu, Wei Lu, Derry Wijaya, Ayu Purwarianti, and Adila~Alfa Krisnadhi (eds.), \emph{Proceedings of the 13th International Joint Conference on Natural Language Processing and the 3rd Conference of the Asia-Pacific Chapter of the Association for Computational Linguistics (Volume 1: Long Papers)}, pp.\  305--329, Nusa Dua, Bali, November 2023. Association for Computational Linguistics.
\newblock \doi{10.18653/v1/2023.ijcnlp-main.20}.
\newblock URL \url{https://aclanthology.org/2023.ijcnlp-main.20/}.

\bibitem[Madaan et~al.(2023)Madaan, Tandon, Gupta, Hallinan, Gao, Wiegreffe, Alon, Dziri, Prabhumoye, Yang, Gupta, Majumder, Hermann, Welleck, Yazdanbakhsh, and Clark]{madaan2023selfrefine}
Aman Madaan, Niket Tandon, Prakhar Gupta, Skyler Hallinan, Luyu Gao, Sarah Wiegreffe, Uri Alon, Nouha Dziri, Shrimai Prabhumoye, Yiming Yang, Shashank Gupta, Bodhisattwa~Prasad Majumder, Katherine Hermann, Sean Welleck, Amir Yazdanbakhsh, and Peter Clark.
\newblock Self-refine: Iterative refinement with self-feedback.
\newblock In \emph{Thirty-seventh Conference on Neural Information Processing Systems}, 2023.
\newblock URL \url{https://openreview.net/forum?id=S37hOerQLB}.

\bibitem[Matthew~Lam et~al.(2024)Matthew~Lam, Thatikonda, and Shareghi]{matthew-lam-etal-2024-closer}
Long~Hei Matthew~Lam, Ramya~Keerthy Thatikonda, and Ehsan Shareghi.
\newblock A closer look at tool-based logical reasoning with {LLM}s: The choice of tool matters.
\newblock In Tim Baldwin, Sergio~Jos{\'e} Rodr{\'i}guez~M{\'e}ndez, and Nicholas Kuo (eds.), \emph{Proceedings of the 22nd Annual Workshop of the Australasian Language Technology Association}, pp.\  41--63, Canberra, Australia, December 2024. Association for Computational Linguistics.
\newblock URL \url{https://aclanthology.org/2024.alta-1.4/}.

\bibitem[Muennighoff et~al.(2025)Muennighoff, Yang, Shi, Li, Fei-Fei, Hajishirzi, Zettlemoyer, Liang, Candès, and Hashimoto]{muennighoff2025s1simpletesttimescaling}
Niklas Muennighoff, Zitong Yang, Weijia Shi, Xiang~Lisa Li, Li~Fei-Fei, Hannaneh Hajishirzi, Luke Zettlemoyer, Percy Liang, Emmanuel Candès, and Tatsunori Hashimoto.
\newblock s1: Simple test-time scaling, 2025.
\newblock URL \url{https://arxiv.org/abs/2501.19393}.

\bibitem[Pan et~al.(2023)Pan, Albalak, Wang, and Wang]{pan-etal-2023-logic}
Liangming Pan, Alon Albalak, Xinyi Wang, and William Wang.
\newblock Logic-{LM}: Empowering large language models with symbolic solvers for faithful logical reasoning.
\newblock In Houda Bouamor, Juan Pino, and Kalika Bali (eds.), \emph{Findings of the Association for Computational Linguistics: EMNLP 2023}, pp.\  3806--3824, Singapore, December 2023. Association for Computational Linguistics.
\newblock \doi{10.18653/v1/2023.findings-emnlp.248}.
\newblock URL \url{https://aclanthology.org/2023.findings-emnlp.248/}.

\bibitem[Parmar et~al.(2025)Parmar, Liu, Goyal, Chen, Le, Mishra, Mobahi, Gu, Wang, Nakhost, Baral, Lee, Pfister, and Palangi]{parmar2025plangenmultiagentframeworkgenerating}
Mihir Parmar, Xin Liu, Palash Goyal, Yanfei Chen, Long Le, Swaroop Mishra, Hossein Mobahi, Jindong Gu, Zifeng Wang, Hootan Nakhost, Chitta Baral, Chen-Yu Lee, Tomas Pfister, and Hamid Palangi.
\newblock Plangen: A multi-agent framework for generating planning and reasoning trajectories for complex problem solving, 2025.
\newblock URL \url{https://arxiv.org/abs/2502.16111}.

\bibitem[Rajani et~al.(2019)Rajani, McCann, Xiong, and Socher]{rajani-etal-2019-explain}
Nazneen~Fatema Rajani, Bryan McCann, Caiming Xiong, and Richard Socher.
\newblock Explain yourself! leveraging language models for commonsense reasoning.
\newblock In Anna Korhonen, David Traum, and Llu{\'i}s M{\`a}rquez (eds.), \emph{Proceedings of the 57th Annual Meeting of the Association for Computational Linguistics}, pp.\  4932--4942, Florence, Italy, July 2019. Association for Computational Linguistics.
\newblock \doi{10.18653/v1/P19-1487}.
\newblock URL \url{https://aclanthology.org/P19-1487/}.

\bibitem[Saparov \& He(2023)Saparov and He]{saparov2023language}
Abulhair Saparov and He~He.
\newblock Language models are greedy reasoners: A systematic formal analysis of chain-of-thought.
\newblock In \emph{The Eleventh International Conference on Learning Representations}, 2023.
\newblock URL \url{https://openreview.net/forum?id=qFVVBzXxR2V}.

\bibitem[Saxton et~al.(2019)Saxton, Grefenstette, Hill, and Kohli]{saxton2018analysing}
David Saxton, Edward Grefenstette, Felix Hill, and Pushmeet Kohli.
\newblock Analysing mathematical reasoning abilities of neural models.
\newblock In \emph{International Conference on Learning Representations}, 2019.
\newblock URL \url{https://openreview.net/forum?id=H1gR5iR5FX}.

\bibitem[Sel et~al.(2024)Sel, Al-Tawaha, Khattar, Jia, and Jin]{10.5555/3692070.3693867}
Bilgehan Sel, Ahmad Al-Tawaha, Vanshaj Khattar, Ruoxi Jia, and Ming Jin.
\newblock Algorithm of thoughts: enhancing exploration of ideas in large language models.
\newblock In \emph{Proceedings of the 41st International Conference on Machine Learning}, ICML'24. JMLR.org, 2024.

\bibitem[Shojaee et~al.(2025)Shojaee, Mirzadeh, Alizadeh, Horton, Bengio, and Farajtabar]{shojaee2025illusionthinkingunderstandingstrengths}
Parshin Shojaee, Iman Mirzadeh, Keivan Alizadeh, Maxwell Horton, Samy Bengio, and Mehrdad Farajtabar.
\newblock The illusion of thinking: Understanding the strengths and limitations of reasoning models via the lens of problem complexity, 2025.
\newblock URL \url{https://arxiv.org/abs/2506.06941}.

\bibitem[Stechly et~al.(2025)Stechly, Valmeekam, and Kambhampati]{stechly2025on}
Kaya Stechly, Karthik Valmeekam, and Subbarao Kambhampati.
\newblock On the self-verification limitations of large language models on reasoning and planning tasks.
\newblock In \emph{The Thirteenth International Conference on Learning Representations}, 2025.
\newblock URL \url{https://openreview.net/forum?id=4O0v4s3IzY}.

\bibitem[Sui et~al.(2025)Sui, Chuang, Wang, Zhang, Zhang, Yuan, Liu, Wen, Zhong, Zou, Chen, and Hu]{sui2025stop}
Yang Sui, Yu-Neng Chuang, Guanchu Wang, Jiamu Zhang, Tianyi Zhang, Jiayi Yuan, Hongyi Liu, Andrew Wen, Shaochen Zhong, Na~Zou, Hanjie Chen, and Xia Hu.
\newblock Stop overthinking: A survey on efficient reasoning for large language models.
\newblock \emph{Transactions on Machine Learning Research}, 2025.
\newblock ISSN 2835-8856.
\newblock URL \url{https://openreview.net/forum?id=HvoG8SxggZ}.

\bibitem[Turpin et~al.(2023)Turpin, Michael, Perez, and Bowman]{10.5555/3666122.3669397}
Miles Turpin, Julian Michael, Ethan Perez, and Samuel~R. Bowman.
\newblock Language models don't always say what they think: unfaithful explanations in chain-of-thought prompting.
\newblock In \emph{Proceedings of the 37th International Conference on Neural Information Processing Systems}, NIPS '23, Red Hook, NY, USA, 2023. Curran Associates Inc.

\bibitem[Valmeekam et~al.(2024)Valmeekam, Stechly, and Kambhampati]{valmeekam2024llmscantplanlrms}
Karthik Valmeekam, Kaya Stechly, and Subbarao Kambhampati.
\newblock Llms still can't plan; can lrms? a preliminary evaluation of openai's o1 on planbench, 2024.
\newblock URL \url{https://arxiv.org/abs/2409.13373}.

\bibitem[Wang et~al.(2023)Wang, Wei, Schuurmans, Le, Chi, Narang, Chowdhery, and Zhou]{wang2023selfconsistency}
Xuezhi Wang, Jason Wei, Dale Schuurmans, Quoc~V Le, Ed~H. Chi, Sharan Narang, Aakanksha Chowdhery, and Denny Zhou.
\newblock Self-consistency improves chain of thought reasoning in language models.
\newblock In \emph{The Eleventh International Conference on Learning Representations}, 2023.
\newblock URL \url{https://openreview.net/forum?id=1PL1NIMMrw}.

\bibitem[Wei et~al.(2022)Wei, Wang, Schuurmans, Bosma, Ichter, Xia, Chi, Le, and Zhou]{10.5555/3600270.3602070}
Jason Wei, Xuezhi Wang, Dale Schuurmans, Maarten Bosma, Brian Ichter, Fei Xia, Ed~H. Chi, Quoc~V. Le, and Denny Zhou.
\newblock Chain-of-thought prompting elicits reasoning in large language models.
\newblock In \emph{Proceedings of the 36th International Conference on Neural Information Processing Systems}, NIPS '22, Red Hook, NY, USA, 2022. Curran Associates Inc.
\newblock ISBN 9781713871088.

\bibitem[Weng et~al.(2023)Weng, Zhu, Xia, Li, He, Liu, Sun, Liu, and Zhao]{weng-etal-2023-large}
Yixuan Weng, Minjun Zhu, Fei Xia, Bin Li, Shizhu He, Shengping Liu, Bin Sun, Kang Liu, and Jun Zhao.
\newblock Large language models are better reasoners with self-verification.
\newblock In Houda Bouamor, Juan Pino, and Kalika Bali (eds.), \emph{Findings of the Association for Computational Linguistics: EMNLP 2023}, pp.\  2550--2575, Singapore, December 2023. Association for Computational Linguistics.
\newblock \doi{10.18653/v1/2023.findings-emnlp.167}.
\newblock URL \url{https://aclanthology.org/2023.findings-emnlp.167/}.

\bibitem[Wu et~al.(2022)Wu, Jiang, Li, Rabe, Staats, Jamnik, and Szegedy]{10.5555/3600270.3602614}
Yuhuai Wu, Albert~Q. Jiang, Wenda Li, Markus~N. Rabe, Charles Staats, Mateja Jamnik, and Christian Szegedy.
\newblock Autoformalization with large language models.
\newblock In \emph{Proceedings of the 36th International Conference on Neural Information Processing Systems}, NIPS '22, Red Hook, NY, USA, 2022. Curran Associates Inc.
\newblock ISBN 9781713871088.

\bibitem[Xie et~al.(2023)Xie, Yu, Zhu, Bai, Gong, and Soh]{xie2023translating}
Yaqi Xie, Chen Yu, Tongyao Zhu, Jinbin Bai, Ze~Gong, and Harold Soh.
\newblock Translating natural language to planning goals with large-language models.
\newblock \emph{arXiv preprint arXiv:2302.05128}, 2023.

\bibitem[Yang et~al.(2023)Yang, Swope, Gu, Chalamala, Song, Yu, Godil, Prenger, and Anandkumar]{10.5555/3666122.3667066}
Kaiyu Yang, Aidan~M. Swope, Alex Gu, Rahul Chalamala, Peiyang Song, Shixing Yu, Saad Godil, Ryan Prenger, and Anima Anandkumar.
\newblock Leandojo: theorem proving with retrieval-augmented language models.
\newblock In \emph{Proceedings of the 37th International Conference on Neural Information Processing Systems}, NIPS '23, Red Hook, NY, USA, 2023. Curran Associates Inc.

\bibitem[Yao et~al.(2023)Yao, Yu, Zhao, Shafran, Griffiths, Cao, and Narasimhan]{10.5555/3666122.3666639}
Shunyu Yao, Dian Yu, Jeffrey Zhao, Izhak Shafran, Thomas~L. Griffiths, Yuan Cao, and Karthik Narasimhan.
\newblock Tree of thoughts: deliberate problem solving with large language models.
\newblock In \emph{Proceedings of the 37th International Conference on Neural Information Processing Systems}, NIPS '23, Red Hook, NY, USA, 2023. Curran Associates Inc.

\bibitem[Ye et~al.(2023)Ye, Chen, Dillig, and Durrett]{10.5555/3666122.3668096}
Xi~Ye, Qiaochu Chen, Isil Dillig, and Greg Durrett.
\newblock Satlm: satisfiability-aided language models using declarative prompting.
\newblock In \emph{Proceedings of the 37th International Conference on Neural Information Processing Systems}, NIPS '23, Red Hook, NY, USA, 2023. Curran Associates Inc.

\bibitem[Zheng et~al.(2024)Zheng, Mishra, Zhang, Chen, Chen, Nova, Hou, Cheng, Le, Chi, and Zhou]{zheng2024naturalplanbenchmarkingllms}
Huaixiu~Steven Zheng, Swaroop Mishra, Hugh Zhang, Xinyun Chen, Minmin Chen, Azade Nova, Le~Hou, Heng-Tze Cheng, Quoc~V. Le, Ed~H. Chi, and Denny Zhou.
\newblock Natural plan: Benchmarking llms on natural language planning.
\newblock 2024.
\newblock URL \url{https://arxiv.org/abs/2406.04520}.

\end{thebibliography}
\bibliographystyle{colm2026_conference}

\appendix
\section{Data and Prompt Examples}
\label{sec:examples}
We use the prompt wordings from \np as is, but provide some examples here for the paper to be self-contained. Here is an example from the calendar scheduling task:
\begin{lstlisting}
You are an expert at scheduling meetings. You are given a few constraints on the existing schedule of each participant, the meeting duration, and possibly some preferences on the meeting time. Note there exists a solution that works with existing schedule of every participant. Here are a few example tasks and solutions:  TASK: You need to schedule a meeting for James and John for one hour between the work hours of 9:00 to 17:00 on Monday.   Here are the existing schedules for everyone during the day:  James has blocked their calendar on Monday during 11:30 to 12:00, 14:30 to 15:00;  John is busy on Monday during 9:30 to 11:00, 11:30 to 12:00, 12:30 to 13:30, 14:30 to 16:30;   Find a time that works for everyone's schedule and constraints. Please provide your solution in a JSON format as as {"start":{"day":"Monday","time"13:30"},
"end":{"day":"Monday","time"14:30"}}.
\end{lstlisting}
To facilitate parsing, we add the last sentence to encourage an output in JSON. Here is an example from the trip planning task:
\begin{lstlisting}
You plan to visit 3 European cities for 7 days in total. You only take direct flights to commute between cities. You want to spend 4 days in Madrid. You would like to visit Dublin for 3 days. You want to spend 2 days in Tallinn. You have to attend a workshop in Tallinn between day 6 and day 7.  Here are the cities that have direct flights: Madrid and Dublin, Dublin and Tallinn.  Find a trip plan of visiting the cities for 7 days by taking direct flights to commute between them. Please provide your solution in a JSON format as as {"itinerary": [{"day_range": "Day 1-2", "place": "Reykjavik"}, {"day_range": "Day 2-4", "place": "Stockholm"}......]}.
\end{lstlisting}
Here is an example from the meeting planning task:
\begin{lstlisting}
You are visiting San Francisco for the day and want to meet as many friends as possible. Solve the problem by considering various different schedules and picking the best one to optimize your goals.  Travel distances (in minutes): Sunset District to Chinatown: 30. Sunset District to Russian Hill: 24. Sunset District to North Beach: 29. Chinatown to Sunset District: 29. Chinatown to Russian Hill: 7. Chinatown to North Beach: 3. Russian Hill to Sunset District: 23. Russian Hill to Chinatown: 9. Russian Hill to North Beach: 5. North Beach to Sunset District: 27. North Beach to Chinatown: 6. North Beach to Russian Hill: 4.  CONSTRAINTS: You arrive at Sunset District at 9:00AM. Anthony will be at Chinatown from 1:15PM to 2:30PM. You'd like to meet Anthony for a minimum of 60 minutes. Rebecca will be at Russian Hill from 7:30PM to 9:15PM. You'd like to meet Rebecca for a minimum of 105 minutes. Melissa will be at North Beach from 8:15AM to 1:30PM. You'd like to meet Melissa for a minimum of 105 minutes. Please provide your solution in a JSON format as as {"itinerary": [{"action": "meet", "location": "Golden Gate Park", "person": "David","start_time": "13:00", "end_time": "14:00"}, ...]}.
\end{lstlisting}
For all methods, we provide one-shot only to illustrate the format of each problem. For \laf, we prompt the LLM to write code which produces the output in the format specified above, though this cannot be guaranteed. We perform simple post-processing with the aid of a small LLM, \texttt{gpt-4.1-mini}. To encourage models' formalization behavior, we additionally append the following wording. 
\begin{lstlisting}
Solve the following problem by
1. First conducting a reasoning to form a generalizible solution method
2. Then applying the formed solution technique in a python script enclosed in ```python and ```.
or
2. Implement a python script enclosed in ```python and ``` that solves the problem using the z3 solver.
\end{lstlisting}

\begin{figure}[!t]
    \centering
    \small
    \begin{tikzpicture}
        \begin{axis}[
            ybar,
            bar width=3pt,
            width=\columnwidth,
            height=4cm,
            symbolic x coords={Human, Llama-3.1-8B, Llama-3.1-70B, DeepSeek-R1-8B, DeepSeek-R1-70B, DeepSeek-R1-671B, gpt-4o-mini, o3-mini},
            xtick=data,
            xticklabel style={rotate=45, anchor=east},
            xticklabels={},
            enlarge x limits=0.1,
            ymin=0,
            ymax=100,
            title={Calendar scheduling},
            title style={yshift=-0.8cm}
        ]
        \addplot coordinates {(Human, 100) (Llama-3.1-8B, 13) (Llama-3.1-70B, 34) (DeepSeek-R1-8B, 5) (DeepSeek-R1-70B, 7) (DeepSeek-R1-671B, 53) (gpt-4o-mini, 39) (o3-mini, 88)};
        \addplot coordinates {(Human, 100) (Llama-3.1-8B, 12) (Llama-3.1-70B, 32) (DeepSeek-R1-8B, 5) (DeepSeek-R1-70B, 8) (DeepSeek-R1-671B, 52) (gpt-4o-mini, 34) (o3-mini, 88)};
        \end{axis}
    \end{tikzpicture}
    \begin{tikzpicture}
        \begin{axis}[
            ybar,
            bar width=3pt,
            width=\columnwidth,
            height=4cm,
            symbolic x coords={Human, Llama-3.1-8B, Llama-3.1-70B, DeepSeek-R1-8B, DeepSeek-R1-70B, DeepSeek-R1-671B, gpt-4o-mini, o3-mini},
            xtick=data,
            xticklabel style={rotate=45, anchor=east},
            xticklabels={},
            enlarge x limits=0.1,
            ymin=0,
            ymax=100,
            title={Trip planning},
            title style={yshift=-0.8cm},
            legend style={at={(0.5,0.5)}, anchor=south,legend columns=-1},
        ]
        \addplot coordinates {(Human, 100) (Llama-3.1-8B, 4) (Llama-3.1-70B, 48) (DeepSeek-R1-8B, 0) (DeepSeek-R1-70B, 30) (DeepSeek-R1-671B, 10) (gpt-4o-mini, 24) (o3-mini, 78)};
        \addplot coordinates {(Human, 100) (Llama-3.1-8B, 4) (Llama-3.1-70B, 49) (DeepSeek-R1-8B, 0) (DeepSeek-R1-70B, 32) (DeepSeek-R1-671B, 71) (gpt-4o-mini, 24) (o3-mini, 59)};
        \legend{1-shot, 5-shot}
        \end{axis}
    \end{tikzpicture}
    \begin{tikzpicture}
        \begin{axis}[
            ybar,
            bar width=3pt,
            width=\columnwidth,
            height=4cm,
            symbolic x coords={Human, Llama-3.1-8B, Llama-3.1-70B, DeepSeek-R1-8B, DeepSeek-R1-70B, DeepSeek-R1-671B, gpt-4o-mini, o3-mini},
            xtick=data,
            xticklabel style={rotate=45, anchor=east},
            enlarge x limits=0.1,
            ymin=0,
            ymax=100,
            title={Meeting planning},
            title style={yshift=-0.8cm}
        ]
        \addplot coordinates {(Human, 100) (Llama-3.1-8B, 0) (Llama-3.1-70B, 77) (DeepSeek-R1-8B, 0) (DeepSeek-R1-70B, 24) (DeepSeek-R1-671B, 0) (gpt-4o-mini, 0) (o3-mini, 0)};
        \addplot coordinates {(Human, 100) (Llama-3.1-8B, 0) (Llama-3.1-70B, 76) (DeepSeek-R1-8B, 0) (DeepSeek-R1-70B, 24) (DeepSeek-R1-671B, 4) (gpt-4o-mini, 72) (o3-mini, 34)};
        \end{axis}
    \end{tikzpicture}
    \caption{The performance of preliminary studies of 1-shot and 5-shot prompting on all 3 domains of \np for \las.}
    \label{fig:few_vs_zero}
\end{figure}

\section{Few-Shot versus One-Shot}
\label{sec:few_vs_zero}
In our preliminary experiment, we compared both 5-shot prompting and 1-shot prompting (default of this paper). For the former, we use the 5-shot prompts provided by \np. We evaluate only generating plan on all 3 tasks, because in-context exemplars for code generation requires expertise in annotation and is thus unrealistic. From Figure~\ref{fig:few_vs_zero}, 5-shot prompting does not increase performance in most cases, while occasionally hurting performance. We inspected its performance gain for some models in meeting planning, and conclude that many errors in 1-shot settings are due to not conforming to a specific output format. These errors are mitigated via 5-shot prompting, while adding little value to insights into reasoning. In later experiments, we relaxed the stringency on output format and rely more on post-processing for a fairer evaluation. 

\section{Python and Z3 Code}
\label{sec:python-z3}
\begin{figure*}[ht!]
  \centering
  \begin{subfigure}[t]{0.48\textwidth}
\begin{lstlisting}[
      language=Python,
      caption={Python solution},
      firstnumber=1
    ]


meeting_duration = 1              
working_start = 9                 
working_end = 17                  
# declare disallowed time
unavailable = [(9, 10)]           

# declare the start time variable
# declare alowed time
for start in range(working_start, working_end - meeting_duration + 1):
    conflict = False              
    # check disallowed time
    for u_start, u_end in unavailable:
        if not (start + meeting_duration <= u_start or start >= u_end):
            conflict = True
            break
    if conflict:
        continue
    # find the earliest time
    print(f"Earliest meeting: {start}:00 to {start + meeting_duration}:00")
    break
\end{lstlisting}
  \end{subfigure}\hfill
  \begin{subfigure}[t]{0.48\textwidth}
\begin{lstlisting}[
      language=Python,
      caption={Z3 solution},
      firstnumber=1
    ]
from z3 import Int, Optimize

meeting_duration = 1
working_start = 9
working_end = 17

# declare the start time variable
start = Int('start')

opt = Optimize()
# declare disallowed time
opt.add(start != 9)             

# declare alowed time
opt.add(start >= working_start)                       
opt.add(start + meeting_duration <= working_end)      

# find the earliest time
opt.minimize(start)

if opt.check() == 0:
    m = opt.model()
    s = m[start].as_long()
    print(f"Earliest meeting: {s}:00 to {s + meeting_duration}:00")
else:
    print("No solution found.")
\end{lstlisting}
  \end{subfigure}
  \caption{Side-by-side comparison of pure Python and Z3-Python code.}
  \label{fig:comparison_python_z3}
\end{figure*}
To showcase the difference between the general Python syntax and the Z3-Python syntax, consider the following example input: 
\begin{lstlisting}
Find the earliest 1-hour meeting time between 9AM-5PM.
9AM-10AM is not available.
\end{lstlisting}
The correct solution (generated by o3-mini) is shown in Figure~\ref{fig:comparison_python_z3}.
Both include two distinctive components of declarative code and algorithmic code.

\section{More Error Analysis}
\label{sec:more_error_analysis}
\begin{figure*}[]
  \centering
\begin{lstlisting}[
      language=Python,
      firstnumber=1
    ]
import json
def compute_itinerary():
    # Input parameters
    total_days = 16
    days_in_lyon = 7
    days_in_bucharest = 7
    days_in_porto = 4
    # Available direct flights:
    # Bucharest <-> Lyon, Lyon <-> Porto
    # Additional constraint:
    # - There is a wedding in Bucharest between day 1 and day 7, so one must be in Bucharest
    #   during that period.
    # - Note: if one flies from a city A to city B on a day, that day counts as a day in both A and B.
    # We have three cities to visit:
    #   Bucharest, Lyon, and Porto.
    # With 2 flights (transitions), the overlapping days count for both cities.
    # Let the flight from Bucharest to Lyon be on day f1 and the flight from Lyon to Porto be on day f2.
    # The days spent in each city are:
    #   Bucharest: days 1 to f1 (inclusive)   -> count = f1
    #   Lyon: day f1 (arrival day) + full days from (f1+1) to (f2-1) + day f2 (departure day)
    #         -> count = 1 + (f2 - f1 - 1) + 1 = f2 - f1 + 1
    #   Porto: day f2 (arrival day) + days (f2+1) to total_days
    #         -> count = 1 + (total_days - f2)
    # We want:
    #   f1 = days_in_bucharest = 7  (so that days 1-7 cover Bucharest,
    #       ensuring the wedding is attended in the first 7 days)
    #   For Lyon: f2 - f1 + 1 = days_in_lyon = 7
    #           => f2 - 7 + 1 = 7  -> f2 = 7 + 6 = 13
    #   Then Porto: 1 + (total_days - f2) = 1 + (16 - 13) = 4 = days_in_porto
    # This gives:
    flight_day_bucharest_to_lyon = 7  # f1
    flight_day_lyon_to_porto = 13      # f2
    # Building the itinerary:
    # The trip plan segments:
    # 1. Bucharest from day 1 to day 7 (day 7 is used for the flight and counts as Bucharest)
    # 2. Lyon from day 7 to day 13 (day 7 from arrival flight, day 13 for departure flight)
    # 3. Porto from day 13 to day 16 (day 13 counts as the arrival day)
    itinerary = [
        {"day_range": "1-7", "place": "Bucharest"},
        {"day_range": "7-13", "place": "Lyon"},
        {"day_range": "13-16", "place": "Porto"}
    ]
    # Output result as a JSON-formatted dictionary
    result = {"itinerary": itinerary}
    return result
def main():
    itinerary_plan = compute_itinerary()
    print(json.dumps(itinerary_plan))
if __name__ == '__main__':
    main()
\end{lstlisting}
  \caption{Another Python program generated by o3-mini on trip planning which hard-codes the solution.}
  \label{fig:hard-coded}
\end{figure*}
Another example of a hard-coded solution due to spurious reasoning is shown in Figure~\ref{fig:hard-coded}.

\section{License}
The \np dataset we use is licensed under Apache License 2.0.

\end{document}